\documentclass[review]{elsarticle} 	
\pdfoutput=1
\usepackage{graphicx}
\usepackage{subfig}
\usepackage{url}
\usepackage[T1]{fontenc}

\usepackage{wrapfig}
\usepackage{amsmath}

\usepackage{xcolor} 

\begin{document}

\newcommand{\etal}{{\it et al.}}
\newcommand{\eg}{{\it e.g.}}

\newcommand{\ngram}{{\it n}-gram }
\newcommand{\ngrams}{{\it n}-grams }

\title{Are metrics measuring what they should? An evaluation of Image Captioning task metrics}

\author[1]{Othón González-Chávez}\ead{ogonzalez@centrogeo.edu.mx}

\author[1,2]{Guillermo Ruiz\corref{cor1}}\ead{lgruiz@centrogeo.edu.mx}

\author[1]{Daniela Moctezuma}\ead{dmoctezuma@centrogeo.edu.mx}

\author[1,2]{Tania Ramirez-delReal}\ead{tramirez@centrogeo.edu.mx}

\cortext[cor1]{Corresponding author}

\affiliation[1]{organization={Centro de Investigación en Ciencias de Información Geoespacial AC (CentroGEO)}, addressline={Circuito Tecnopolo Norte}, city={Aguascalientes}, postcode={20313}, country={Mexico}}

\affiliation[2]{organization={Consejo Nacional de Ciencia y Tecnología (CONACyT)-Centro de Investigación en Ciencias de Información Geoespacial AC (CentroGEO)}, addressline={Circuito Tecnopolo Norte}, city={Aguascalientes}, postcode={20313}, , country={Mexico}}

\begin{abstract}Image Captioning is a current research task to describe the image content using the objects and their relationships in the scene. To tackle this task, two important research areas converge, artificial vision, and natural language processing. 
In Image Captioning, as in any computational intelligence task, the performance metrics are crucial for knowing how well (or bad) a method performs.
In recent years, it has been observed that classical metrics based on $n$-grams are insufficient to capture the semantics and the critical meaning to describe the content in an image. Looking to measure how well or not the set of current and more recent metrics are doing, in this article, we present an evaluation of several kinds of Image Captioning metrics and a comparison between them using the well-known MS COCO dataset. 
The metrics were selected from the most used in prior works, they are those based on $n$-grams as BLEU, SacreBLEU, METEOR, ROGUE-L, CIDEr, SPICE, and those based on embeddings, such as BERTScore and CLIPScore.
For this, we designed two scenarios; 1) a set of artificially build captions with several qualities, and 2) a comparison of some state-of-the-art Image Captioning methods. Interesting findings were found trying to answer the questions: Are the current metrics helping to produce high-quality captions? How do actual metrics compare to each other? What are the metrics \emph{really} measuring?
\end{abstract}


\maketitle

\section{Introduction}

The task of Image Captioning (IC) has been one of the most studied in recent years due to its relevance in many practical applications. Also, IC task can be used as a base for other natural language generation challenges like Question Answering, Video Captioning, and smart video surveillance applications, among others. 

Its primary purpose is to describe a scene using natural language, describing the image content, the objects on it, and the relationship among them. Throughout the years, the evolution of the techniques used for the proposed models has been remarkable; they have gone from probabilistic models to deep learning procedures like Convolutional Neural Networks (CNN), Long Short-Term Memory (LSTM), and recently, transformers. 

Undoubtedly, the use of transformers has improved the quality of the captions produced by the research community.

The data used to evaluate the proposed methods is also a crucial aspect; its amount and quality during training significantly impact the test results. Not many years ago, datasets had less than 100k examples; nowadays, the best models have trained on huge collections bigger than 100 million elements \cite{zhang2021vinvl, hu2021scaling}. Not only has their size changed, but the number of captions per image and the diversity of the pictures increased as well.

This increase in the datasets' sizes has led to the creation of bigger models. In fact, it is not strange to see systems with more than 100 million parameters; some might have more \cite{hu2021scaling}. As a consequence, the training time has also increased significantly, some models would need seven years to learn their weights if they were trained on common computers \cite{dosovitskiy2021an}. It leaves most research labs as simple spectators as they do not have the resources to replicate such models or modify them to achieve some improvement. Fortunately, some outstanding small models are very close in quality \cite{cornia2020m2} which implies that they are much more efficient.

Given the prior context, observing all this model's evolution and improvements, there is an important part that has been maintained almost unchanged through all these years, \emph{the metrics} used for evaluating such models. These evaluation metrics must rate how good the generated captions are, comparing the system's output with the human-generated data. Maybe, a perfect way to evaluate the generated captions is to give them to human judges to determine their quality, but this would be slow and expensive.

The aforementioned reasons force the scientific community to seek for another way to \emph{score} their models and compare them, like building a \emph{reliable} automatic metric (or set of metrics) to score the quality of the generated captions. To do that, we currently have a set of most used performance metrics, like Bilingual Evaluation Understudy (BLEU) \cite{Papineni02bleu:a}, Recall-Oriented Understudy for Gisting Evaluation (ROUGE) \cite{lin-och-2004-orange}, Metric for Evaluation of Translation with Explicit ORdering (METEOR) \cite{Lavie2007}, Consensus-based Image Description Evaluation (CIDEr) \cite{vedantam2015cider}, and Semantic Propositional Image Caption Evaluation (SPICE) \cite{anderson2016spice}. 

They all rely on counting the number matching of $n$-grams present in the proposed caption against the reference or human-generated annotations, only with some modifications. It is interesting to note that only CIDEr and SPICE were specifically designed for the task of IC; the remaining were designed for a different task like machine translation but applied to IC evaluation successfully. 

However, only having a set of metrics is not enough; seeing how they work is also essential. Unfortunately, actual most-used metrics, based on $n$-grams, penalize creativity and diversity, usually produced in semantic aspect, so we must evaluate and improve them to build better alternatives.

Some efforts have been undertaken in this sense; Sai et al. \cite{sai2022survey} propose a survey of metrics evaluation in some natural language generation systems, not only for the IC task. Also, Sai et al. presented a very useful taxonomy for existing evaluation metrics. Furthermore, some efforts for metrics evaluation and metrics proposal have been done (see \cite{kilickaya2016re} and \cite{tanti2018pre}). In \cite{elliott2014comparing} a work is presented utilizing a correlation coefficient to find the best metric between humans and automatic sentences generated by Image Captioning models. Nevertheless, sometimes the score obtained does not necessarily show a good caption describing a scene; it only denotes the similarity between the ground truth sentence and the generated one. 
For example the sentences ``a kid is playing in the park with a kite'' and ``a dog is playing in the park with a ball'' have a high BLEU score but come from very different pictures.

Knowing how metrics perform is crucial to improve any task solutions or new models. Looking to get some insights into the performance of IC metrics, in this article, we analyze the behavior of several current and most-used IC metrics such as those based on \ngrams  (BLEU, SacreBLEU, METEOR, ROGUE-L, CIDEr, and SPICE), and those based on embeddings (BERTScore and CLIPScore).

We compare them in a quantitative and methodological way. Two experimental scenarios were designed, 1) a set of artificially build captions having several quality levels, and 2) compared with some state-of-the-art methods. All the experiments were designed to measure the performance of the most common metrics under different scenarios.
Finally, we present strong evidence for the need to transition to better metrics than the ones mostly used by the IC community nowadays, based on both previous studies and the results of our experiments.


This article is organized as follows.
In Section \ref{sec:image-caption} some related works from classical to newest and outstanding are described.
In Section \ref{sec:metrics} we review the most common metrics used by the research community and some of the new generation of metrics based on Deep Learning. Also, we present some efforts with findings related to the need to improve the metrics. Finally, in Section \ref{sec:ranking}, we present our own experiments in the MS-COCO dataset with interesting conclusions. Section \ref{sec:conclusions} presents 
some conclusions and recommendations.


\section{Related work}
\label{sec:image-caption}

As already stated, the task of IC is the process of automatically generating a text description of an image. The description should contain the most critical elements depicted in the image but also their interactions, just as a human would write a caption. This is a complex task because there can be multiple valid captions for the same picture. Much research has been done to solve IC in recent years; some of the first attempts used templates to output a caption \cite{farhadi2010every, 10.5555/2380816.2380907}, or select the description that best fitted the image from a set of fixed captions \cite{Gong2014ImprovingIE}. This kind of model usually failed to generalize.

Nowadays, the most outstanding approaches for dealing with the IC task are those based on Deep Learning (DL). 
Some early models worth to mention are \cite{pmlr-v32-kiros14, DBLP:journals/corr/MaoXYWY14a,7298856}.

The caption is usually produced by a Recurrent Neural Network (RNN) which predicts the probability distribution of a sequence of words. To do that is necessary a model called Language Model.
In this regard, Vinyals et al. showed in \cite{showandtell} the Neural Image Caption Generator (NIC) model, where the image features were extracted from a CNN and then were put as the beginning state from an LSTM \cite{10.1162/neco.1997.9.8.1735} getting acceptable results. Later, many DL methods took the image features from a CNN and then, with a Language Model (LM), created the caption \cite{10.1145/3115432}. 
This basic recipe has been changing over the years with some additional ideas; some of the most important are the Attention Mechanism, The Transformer, and the inclusion of Reinforcement Learning. 
The Attention \cite{doi:10.1080/135062800394667} mechanism can be used to allow each of the words to have access to the image features and to focus on certain areas. Some examples of models using attention mechanism could be found in~\cite{69e088c8129341ac89810907fe6b1bfe, 10.5555/3298023.3298174, Anderson_2018_CVPR, pmlr-v37-xuc15, pedersoli:hal-01428963,7780872,8100150, ke2019reflective}.

The Reinforcement Learning involves the interactions of an actor in an environment that has to take an action to achieve a goal. In IC, the actor is the model, the environment is the partially produced caption plus the input image, the action is the next chosen word, and the reward is usually given by the CIDEr metric \cite{8099614}. This type of learning helps to reduce the so called \textit{exposure bias}~\cite{DBLP:journals/corr/RanzatoCAZ15}. Some of the first examples of IC models using Reinforcement Learning are \cite{ren2017deep, 8099614, Zhang-NIPS}.


In \cite{vaswani2017attention}, Vaswani et al. proposed the Transformer architecture, a many-to-many neural network that relies on attention and was used for machine translation. It contains an \textbf{encoder} for feature extraction of the input and a \textbf{decoder} that pays attention to the full encoded sentence and the previous words in the translated sentences. 

This encoder-decoder structure was found to be very effective in different tasks like language modeling \cite{Solaiman2019ReleaseSA, devlin-etal-2019-bert}, text and image classification \cite{NEURIPS2019_dc6a7e65, 10.1007/978-3-030-32381-3_16}, and image and video retrieval \cite{song-pvse-cvpr19}. After the transformer introduction, many image caption models used this architecture with great success, some examples are \cite{cornia2020m2, pan2020x, dosovitskiy2021an, Liu2021CPTRFT, li2020oscar, zhang2021vinvl,hu2021scaling,desai2021virtex}.

Nevertheless, a downside of using a transformer for IC is the data needed during training. We can see an increment in the size of the datasets concerning the models using an RNN. For example, Zhang et al. \cite{zhang2021vinvl} used 8.85 million images to train their model. The considerable size of the dataset and the very long training times make it impossible to replicate their models without the appropriate hardware.



\begin{table}
\centering
\begin{tabular}{l c c c}
\hline
 Method & Types/Year & Data & Metrics \\
 \hline
  & 2010-2014  \\
 \hline
 Farhadi \cite{farhadi2010every}  & SL & S & B\\
 Mitchell \cite{10.5555/2380816.2380907} & SL & M & H\\
 Gong \cite{Gong2014ImprovingIE} & SL & S & -\\

 Kiros \cite{pmlr-v32-kiros14} & MS, CNN, SL & S, M & B \\
  \hline
  & 2015  \\
 \hline
 Mao \cite{DBLP:journals/corr/MaoXYWY14a} & MS, RNN, CNN, SL & S & B, M, R, C\\
 Chen \cite{7298856} & MS, RNN, CNN, SL & S & B, M, C\\
 Xu \cite{pmlr-v37-xuc15} & VS+TS, RNN+A, CNN, SL & S & B, M \\
 Liu \cite{pmlr-v37-xuc15} & VS+TS, RNN+A, CNN, SL & S & B, M\\
 Fang \cite{fang2015captions} & MS, CNN, SL& S & B, M, R, C, H\\
 Vinyals \cite{showandtell} & VS+TS, RNN, CNN, SL & S, M* & B, M, C, H\\
 \hline
  & 2016  \\
 \hline
 Wang \cite{10.1145/3115432}& MS, RNN, CNN, SL & S & B, M, C\\
 Yang \cite{NIPS2016_9996535e} & MS, RNN+A, CNN, SL & M & B, M, R, C\\
 You \cite{7780872} & VS+TS, RNN+A, CNN, SL & S, M & B, M, R, C\\
 Tran \cite{Tran2016RichIC} & MS, CNN, SL & S & H \\
 Mao \cite{7780378} & VS+TS, RNN, CNN, SL,& S & H\\
 Sugano \cite{Sugano2016SeeingWH} & VS+TS, RNN+A, CNN, SL & S & B, M, R, C\\
 \hline
 & 2017  \\
 \hline
 Pedersoli \cite{pedersoli:hal-01428963}& MS, RNN+A, CNN, SL & S & B, M, C\\

 Chen \cite{8100150} & VS+TS, RNN+A, CNN, SL & S & B, M, R, C\\
 
 Ke \cite{ke2019reflective} & VS+TS, RNN+A, CNN, SL & M & B, M, R, C, S \\
 Ren \cite{ren2017deep} & MS, RNN, CNN, RL & S & B, M, R, C \\ 
 Rennie \cite{8099614} & MS, RNN+A, CNN, RL & S & B, M, R, C\\
 Zhang \cite{Zhang-NIPS} &  VS+TS, RNN, CNN, RL & S & B, M, R, C\\
  Karpathy \cite{10.1109/TPAMI.2016.2598339} & MS, RNN, CNN, SL & S, M & B, M, C\\
 \hline
  & 2018-2020  \\
 \hline
 Anderson \cite{Anderson_2018_CVPR} & VS+TS, RNN+A, CNN, SL & M & B, M, R, C, S\\ 
 Huang \cite{huang2019attention} & VS+TS, RNN+A, CNN, SL, RL & M & B, M, R, C, S \\
 
 Cornia \cite{cornia2020m2} & VS+TS, TR, CNN, SL, RL & M & B, M, R, C, S \\
 Pan \cite{pan2020x} & MS, TR, CNN, SL, RL & M & B, M, R, C, S \\
 Wang \cite{wang2020visual} & VS+TS, RNN+A, CNN, NSL, SL & M & B, M, R, C, S \\
 Li \cite{li2020oscar} & MS, TR, CNN, PT, SL & L & B, M, C, S \\

 \hline
 & 2021  \\
 \hline
 Liu \cite{Liu2021CPTRFT} & VS+TS, TR, V-TR, SL, RL & M & B, M, R, C\\
 Zhang \cite{zhang2021vinvl} & MS, TR, CNN, PT, SL RL & L & B, M, C, S \\
 Hu \cite{hu2021scaling} & MS, TR, CNN, PT, SL, RL & L & B, M, C, S \\
 Desai \cite{desai2021virtex} & MS, TR, CNN, SL & M & C, S\\
 Galatolo \cite{galatolo2021generating} & MS, TR, NSL, & L & H\\
 
 \hline
\end{tabular}

\caption{Chronological overview of Image Captioning methods. The meaning of the acronyms for the Type column VS+TS: Visual Space + Text Space, MS: Multimodal Space, RNN: Recurrent Naural Network, RNN+A: RNN + Attention, TR: Transformer, CNN: Convolutional Neural Network, V-TR: Visual Transformer, SL: Supervised Learning, NSL: Non-Supervised Learning, RL: Reinforcement Learning, PT: pre-training. Acronyms for the Data S: Small ($<100$K), M: Medium ($<1$M), L: Large. For the metrics B: Bleu, M: Meteor, R: Rouge, C: CIDER, S: Spice, H: Human} \label{tab:metrics}
\end{table}

\section{The metrics} \label{sec:metrics}

 Since image analysis became a discipline, efforts had been done to understand on a global level what is happening in a certain image or scene. The ultimate goal of being able to tell the context in an understandable and reliable manner is yet to be accomplished.
 Due to the increasing development of proposals to deal with this task, the important question that arises is \emph{``how do we measure their progress?"}, which is a non-trivial task.
 
 In this section, we will focus on the different techniques used for scoring the IC performance and how they have evolved through time, also a comparison between them under certain scenarios is presented. 
 First, we will describe the most known and used metrics depicting milestones achieved by the state-of-the-art (SoTA) IC techniques and how they were measured or scored.
 Secondly, we propose new ideas for scoring the current metrics for IC task, and finally we will try to foresee what efforts can be made to further advance this area of machine learning (ML).
 
 As an initial point, and as a global summary, Table \ref{tab:metrics} shows a compendium of IC solutions, their type of methods used, the data employed, also the metrics used to evaluate the performance. As stated before, the main objective of the IC task is to generate an accurate or representative set of ordered words that describe the content depicted on an image. One can make an abstraction of this task to the {\it translation} of one language (the image features generated by certain algorithm space) to another language (the annotation or, ultimately, the natural language space). With this abstraction we can then apply the same comparison scores used to assess the Natural Language Processing (NLP) translation algorithms or models to score their Image Captioning counterparts. It is then expected that some of the scoring tools are taken from its translation techniques counterparts. 
 
 The popular metrics have received criticism and some studies do not correlate them well with human judgments, which is a very important issue. These metrics will be described in the next section, and they were organized into two principal categories, namely \ngram based, and embedding based.

\subsection{Metrics based on \ngram scoring} 
\label{sec:ngram}
As stated, first ML captioning models were focused on {\it translating} the image features domain onto tags or, ideally, sequences of words. Once we have a sequence it can be compared to a human-generated caption to score its performance. The question remains, how can we compare adequately two sets of words? First attempts were made using \ngrams\footnote{An \ngram is a sub-sequence of $n$ elements, taken from a larger sequence.} as the minimum part of a sequence of words, where the $n$ stands for the number of such minimum parts, i.e., monograms, bigrams, trigrams, tetragrams\footnote{Most \ngram based scores use the tetragram as the largest subpart of a sentence.}, and so on.

\subsubsection{BLEU and SacreBLEU}\label{sec:bleu}

One of the first metrics to arrive was the BLEU (Bilingual Evaluation Understudy) score \cite{Papineni02bleu:a}, still used to assess many modern models. It looks for {\it n}-grams on the reference caption that repeat in the proposed caption. As it is proposed on the original paper, BLEU seeks for {\it shared} words in the candidate sentence that are repeated on the reference sentence. Furthermore, we need to address that machine translation relies in tokenization of words and not on the strings of characters that define a word. Therefore, it is worth to mention, that the \ngram focuses on exact instances of the sequence of tokens, and it will also depend largely on the tokenization technique used to encode the sentence. Moreover, it will look only for content, rather than context or semantics in the target sequence.

The BLEU score is defined as

\begin{equation}
    \textnormal{BLEU}_N=\textnormal{BP}\cdot \exp\left(\sum_{n=1}^N w_n \log p_n\right)\, ,
\end{equation}

\noindent where BP is a brevity penalty applied to the candidate sentence, $w_n$ is a weight for the average of the BLEU score which must sum to 1, and $p_n$ is the precision score obtained from counting the \ngram matches of the candidate tokenized phrase in respect to the references.

We must highlight that the authors of \cite{Papineni02bleu:a} had stated that BLEU is better suited for large corpora of text, which in IC is rarely the case. Moreover, the existence of non-shared $n$-grams could lead to zeroes on the $\textnormal{BLEU}_N$ score in several cases of a dataset, plummeting the final overall score on large curated datasets, such as MS-COCO, Flicker-8k, Imagenet, etc.

Early problems were found in using BLEU as a score. In 2006, Callison-Burch et al. \cite{callison-burch-etal-2006-evaluating} conducted an analysis on how BLEU can lead to a wrong judgment on a set of sentences. They showed that a sentence can have a lot of permutations (like $10^{73}$) which will produce the exact same BLEU score. Also, they presented how to take a combination of the \ngram from different references to get a sentence with a high score, with captions that made no sense for a human reader.

In the original paper, the BLEU score had different parameters that could be tweaked to adjust the sensibility of the score to different NLP behaviors. It was discovered, in a later meta-analysis that different techniques used different parameters to report their results. 

A proposal by Matt Post in~\cite{post-2018-call} called for a consensus in the parameterization and tokenization of the techniques used while calculating BLEU used to report new models. He found several discrepancies between results reported on different models using different parameters and tokenization for the BLEU score. The result was the arrival of SacreBLEU, which is now the most used BLEU score version for reporting results in NLP models. One of the key introductions of SacreBLEU is that it expects detokenized references, to later apply its own standard tokenization method which cannot be tweaked by the user. It also integrates a tool that resumes the parameters used by the script, including, for example, the type of tokenization used, the number of references supplied, the version of SacreBLEU run, etc. This will ultimately give the reader of a model report a more accurate understanding of how the metric was applied to it.

\subsubsection{METEOR}

It is another \ngram based metric, described in \cite{Lavie2007}, and considered as a classic metric, which is often used by many authors for reporting their model's behavior. A {\it one-to-one} mapping of words is done before the actual METEOR score is calculated and also Porter stemming is applied. Possible synonyms are also taken into account using Wordnet's {\it synsets} \cite{beckwith1990wordnet}. It is worth mentioning that METEOR, adversely to what BLEU does, only takes into account the best scoring reference, if many are supplied, for its calculation.

METEOR obtains correspondent precision and recall, denoted as $P$ and $R$, respectively, and then applies the parameterized harmonic mean \cite{van1979information} as follows,

\begin{equation}
    F_{\textnormal{mean}}=\frac{P\cdot R}{\alpha\cdot P+(1-\alpha)\cdot R}\,\,\,\,\,\, ,
\end{equation}

\noindent then it incorporates a penalty, $\textnormal{Pen}=\gamma\cdot \textnormal{frag}^\beta$, where $\alpha,\, \beta$, and $\gamma$ are tunable parameters and frag is a fragmentation fraction taking into account the maximum possible chunks where unigrams are adjacent, and the number of matches between the best reference and the candidate. Finally, the score is obtained as 

\begin{equation}
    \textnormal{METEOR}=(1-\textnormal{Pen}) \cdot F_{\textnormal{mean}}\,\,\, .
\end{equation}

It is worth mentioning that, similarly to BLEU, METEOR is subject to parameterization tuning, namely the $\alpha,\, \beta$, and $\gamma$ values. In the original paper \cite{Lavie2007}, different tuned values are {\it suggested} to better correlate with human judgment for translation between different languages, but the {\it elephant-in-the-room} remains present, as in BLUE score case, where no standard is yet presented for the IC problem.

\subsubsection{ROUGE-L}

This metric, as presented in its original paper \cite{lin-och-2004-orange}, is a summary evaluation metric, based on \ngram occurrences. From the family of ROUGE metrics, ROUGE-L, or Longest Common Sequence (LCS) based ROUGE, is the preferred metric when dealing with IC task. ROUGE uses sequences of words, or tokens to be more precise. ROUGE-L uses the F-measure of the LCS to obtain a measure of the similarity of the sentences $X$ and $Y$, as follows

\begin{equation}
    R_{lcs}=\frac{\textnormal{LCS}(X,Y)}{m}\,\,,
\end{equation}

\begin{equation}
    P_{lcs}=\frac{\textnormal{LCS}(X,Y)}{n}\,\,,
\end{equation}

\begin{equation}
    F_{lcs}=\frac{(1+\beta^2)R_{lcs}P_{lcs}}{R_{lcs}+\beta^2 P_{lcs}}\,\,,
\end{equation}

\vspace{0.5cm}

\noindent where $m$ and $n$ are the length of $X$ and $Y$, respectively, and $\beta= P_{lcs}/R_{lcs}$.

\subsubsection{CIDEr}

Consensus-based Image Description Evaluation \cite{vedantam2015cider} (CIDEr), is a metric that arose in the search for a metric that could better correlate with human judgment, by means of acquiring a {\it consensus} between the several proposed captions, or references, in a dataset. It has become a quality measure since it looks for similarity between several human-annotated captions. The authors created a ground truth of triplets that contained the more similar sentence from a given pair. The accuracy of the metric was calculated by comparing the scores given by CIDEr with the ground truth.

First, a stemming process i applied to all the words. Each sentence is then represented by its \ngrams which are weighted using Term Frequency Inverse Document Frequency (TF-IDF) \cite{SprckJones2004ASI}. $s_{ij}$ is the $j$th reference  sentence of image $I_i$. The weight of the \ngram $k$ on the $j$th reference is denoted $g_k(s_{ij})$ and $g_k(c_i)$ for the candidate caption $c_i$.

CIDEr is calculated as follows:

\begin{equation}
    \mathrm{CIDER}_{n}\left ( c_{i},S_{i} \right )=\frac{1}{m}\sum_{j}\frac{\mathbf{g^{n}}\left ( c_{i} \right )\cdot \mathbf{g^{n}}\left ( s_{ij} \right )}{\left \|  \mathbf{g^{n}}\left ( c_{i} \right ) \right \|\left \|\mathbf{g^{n}}\left ( s_{ij} \right )  \right \|},
\end{equation}

\noindent where $n$ is the length of the $n$-gram and $\mathbf {g^n}(c_i)$ is a vector constructed by taking all the $g_k(c_i)$ (see \cite{vedantam2015cider} for details). Therefore, $\|\mathbf {g^n}(c_i)\|$ is the vector's magnitude, and similar symbology applies for $\mathbf {g^n}(s_{ij})$. Finally, the overall $\textnormal{CIDER}(c_i,S_i)$ is computed by taking a weighted sum of all $\sum_{n=1}^N\omega_n\textnormal{CIDER}_n(c_i,S_i)$. Conventionally $\omega_n$ is uniform, and $N=4$.

\subsubsection{SPICE}

The problem with the need of $n$-gram overlapping to achieve high scores in the classical metrics was already outlined by Anderson {\it et al.} in \cite{anderson2016spice}, proposing a Semantic Propositional Image Caption Evaluation (SPICE). By transforming the reference and candidate sentence into what they call a {\it scene graph}, SPICE aim to convey relationships between objects and their characteristics, their relative positions or actions being held between them. To achieve this they use a pre-trained dependency model based on the Stanford Scene Graph Parser \cite{schuster2015generating}, but tuned to suit the task.

So the SPICE score is defined as the $F$-score obtained by comparing the two graphs as follows

\begin{equation}
    P(c,S)=\frac{\mid T(G(c))\otimes T(G(S))\mid}{\mid T(G(c)\mid}\, ,
\end{equation}

\begin{equation}
    R(c,S)=\frac{\mid T(G(c))\otimes T(G(S))\mid}{\mid T(G(S)\mid}\, ,
\end{equation}

\begin{equation}
    \textnormal{SPICE}(c,S)=F_1(c,S)=\frac{2\cdot P(c,S)\cdot R(c,S)}{P(c,S)+ R(c,S)}\, ,
\end{equation}
\vspace{0.3cm}

\noindent where $\otimes$ is a binary matching operator that returns matching tuples in two scene graphs, $T$ is a function that return logical tuples of a scene graph, $G$ is the scene graph obtained in the parsing step, $P$ and $R$ are precision and recall respectively, $c$ is the candidate sentence, and $S$ is the set of references.

\subsection{Embedding based metrics} \label{sec:embedding}

With the advent of Deep Learning techniques, it became clear that NLP will require some more refined scoring techniques in order to remain relevant to the SoTA and relevant to compare these trending techniques. Techniques like Word2Vec \cite{mikolov2013distributed, mikolov2013efficient}, and most recently NLP models like BERT, quickly showed that words, or meaning, can be embedded into upper dimensional spaces representations ($\sim\!\!\!300$ for Word2Vec and up to $\sim\!\!\!700$ for BERT) and be treated as vectorial representations of their character string counterparts. 

These new vector representations of meaning could be feeded to Neural Networks trained specifically to treat these new representations in ways that $n$-grams cannot achieve. Furthermore, vectorial representations of words can be clustered in these new spaces and therefore will be \textit{close} to each other if their meaning is similar, and apart if their meaning is not. There is a family of metrics in which these features are taken as an advantage, and vectorial operations are applied to the embeddings to calculate their similarity. 

\subsubsection{BERTScore}

For the vector representations of two different words ($\mathbf {A}$ and $\mathbf {B}$), the cosine similarity, is computed using the dot product divided by the magnitude of the vectors, and its value is given by

\begin{equation}
\mathrm{cosine\, similarity}=\frac{\mathbf{A}\cdot \mathbf{B}}{\left \| \mathbf{A} \right \|\cdot \left \| \mathbf{B} \right \|}=\frac{\sum_{i=1}^{n}A_iB_i}{\sqrt{\sum_{i=1}^{n}A_{i}^{2}}\sqrt{\sum_{i=1}^{n}B_{i}^{2}}},
\end{equation}

\noindent where $n$ is the dimensionality of the embedding space.

Finally, the BERTScore \cite{zhang2019bertscore} is calculated by taking the maximum score, {\it i.e.} greedy matching, of all the candidate tokens and calculating the precision, recall, and F1 scores. An optional importance weight can be taken into account for some tokens if the user finds it useful.

\begin{figure}
  \centering
  \subfloat[]{\includegraphics[height=2in]{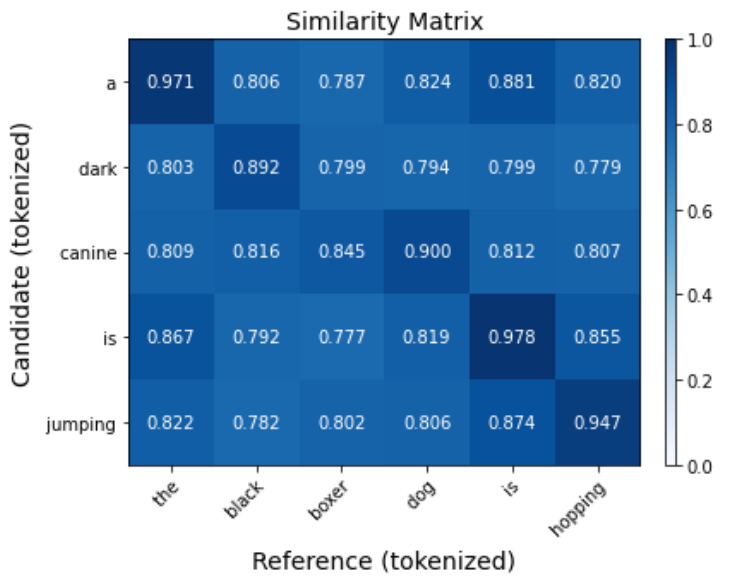}\label{fig:bert1}}
  \hfill
  \subfloat[]{\includegraphics[height=2in]{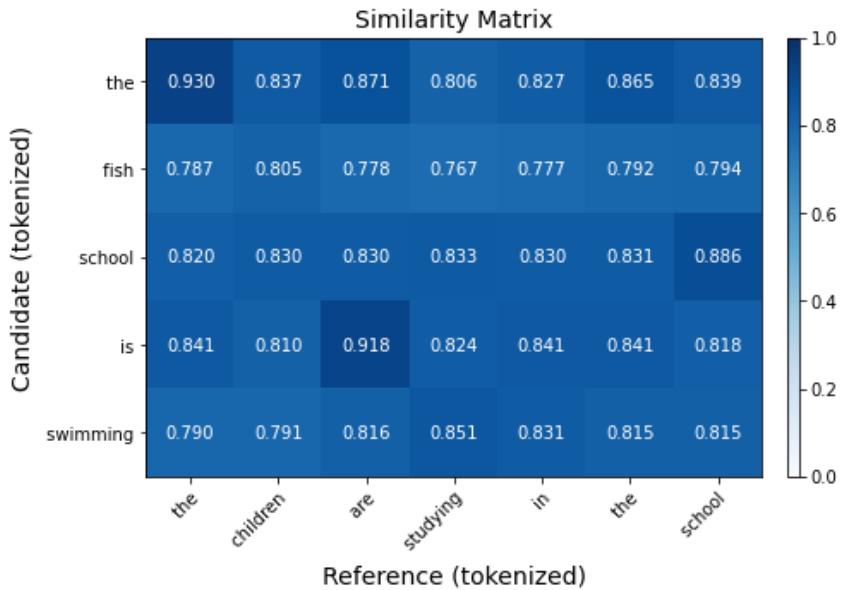}\label{fig:bert2}}
  \caption{Cosine similarity matrix for two example references and candidate sentences using BERT embeddings. In (a) note that even though {\it canine} and {\it dog} are not the same word, their embeddings are close in the higher dimensional space, the same happens with {\it jumping} and {\it hopping}. In (b) the word {\it school} have different meanings on each sentence, therefore their cosine similarity is not as high.\label{fig:bert_comp}}
\end{figure}

It is worth mentioning that, since the embedding is made using the BERT method, the whole sentence is used on the embedding of the word, hence it takes into account the context of the word. For example, the word {\it train} will be differently embedded if it is used in the sentence, {\it The train is passing through the bridge}, than when it is used in the sentence, {\it The runner must train hard to win the competition}. Also, words like canine, dog, and pet, will have a high cosine similarity and therefore the words embedded in the same context should remain close also. Some examples are presented in Figure \ref{fig:bert_comp} where one can see some high similarities between words like black and dark; dog and canine; and much less similar when comparing dog and dark; hopping and dark.

\subsubsection{CLIPScore}
In \cite{DBLP:journals/corr/abs-2103-00020} Radford et al. made public a model trained on images and their descriptions taken from public sites on the internet using contrastive learning called CLIP. This model uses two encoders, one for the image and one for the description, and learns to embed both to similar vectors and at the same time, tries to tell apart an image from the other descriptions. CLIP was trained on a dataset of 400M images with descriptions taken from the internet. The goal was to create a \emph{zero-shot} model with the ability to classify images using a variety of datasets. Quickly after its release, CLIP has been used in plenty of tasks like image and video retrieval, image generation, among others.

Hessel et al. used CLIP as a metric in \cite{Hessel2021CLIPScoreAR}. They used the trained CLIP to measure the similarity of an image and a caption without the need for any references, they called it CLIPScore. In the case the references were available, they also measured the similarity of the predicted caption with them and took the harmonic mean to get CLIPScore$^{\textnormal{ref}}$. In their experiments, this metric was found to correlate stronger with human judgment than the other metrics.

\subsection{Metrics Examples}\label{sec:examples}

To give the reader a concrete example of the captions that can be obtained in a real caption task, we took an example from the actual MS-COCO dataset, namely the picture with ID 544 and its set of reference captions. We here present the values obtained for each of the metrics described previously.
\vspace{0.5cm}

Let us define the set of the reference sentences $\mathcal{R}$ as follows:
\vspace*{2mm}

\begin{equation*}
\mathcal{R}=\left\{
\begin{aligned}
 &\text{\footnotesize `A man swinging a bat at a baseball on a field.'} \\
 &\text{\footnotesize `A man that has a baseball bat standing in the dirt.'} \\
  &\text{\footnotesize `The stands are packed as a baseball player} \\
  & \text{\footnotesize     in a gray uniform holds a bat as a  catcher holds out his mitt.'} \\
 &\text{\footnotesize `A baseball player holding a bat over the top of a base.'}\\
 &\text{\footnotesize `A baseball game is going on for the crowd.'}
\end{aligned} \right.
\end{equation*}

\vspace*{3mm}

\noindent and let us take a candidate sentence such as $c=$ \{`A baseball player is swinging his bat to hit the ball.'\}, we can then obtain the respective BLEU scores.  The range of values are from 0 (worst) to 1 (best). The BLEU results are:

\begin{equation*}
    \begin{aligned}
     \text{BLEU}_1(c,\mathcal{R})&= 0.727272\\
     \text{BLEU}_2(c,\mathcal{R})&= 0.381385\\
     \text{BLEU}_3(c,\mathcal{R})&= 0.252829\\
     \text{BLEU}_4(c,\mathcal{R})&= 0.000003\\
    \end{aligned}
\end{equation*}
Note how big is the range of values between the different BLEU scores. Despite the candidate caption $c$ is a high quality description of the image, the $\text{BLEU}_4$ is practically zero because of the order of the words. 

We can continue, hence, applying the rest of the presented metrics with the goal to see how they perform between each other. The best reported results for SoTA models for METEOR are in the range of $0.30$, for ROUGE-L $0.60$, CIDEr $1.40$ and SPICE $0.25$ \cite{Stefanini2021}. 

Following with the example, its results are:

\begin{equation*}
    \begin{aligned}
     \text{METEOR}(c,\mathcal{R})&= 0.215727\\
     \text{ROUGE-L}(c,\mathcal{R})&= 0.448529\\
     \text{CIDEr}(c,\mathcal{R})&= 1.048888\\
     \text{SPICE}(c,\mathcal{R})&= 0.222222\\
    \end{aligned}
\end{equation*}

All of the aforementioned scores were calculated using the \texttt{pycocoevalcap} library\footnote{\url{https://pypi.org/project/pycocoevalcap/}}. We will now use the \texttt{datasets}\footnote{\url{https://huggingface.co/docs/datasets/index}} project from \texttt{huggingface} to determine the BERTScore and SacreBLEU metrics.
The best results reported for SacreBLEU are around $0.40$. Best possible result for BERTScore is $1.00$, but SoTA models are obtaining values around $0.94$ \cite{Stefanini2021}.

\begin{equation*}
\begin{aligned}
    \text{SacreBLEU}(c,\mathcal{R})=0.165903\\
    \text{BERTScore}(c,\mathcal{R})=0.943384
\end{aligned}
\end{equation*}

\begin{figure}
    \centering
    \includegraphics[width=0.5\textwidth]{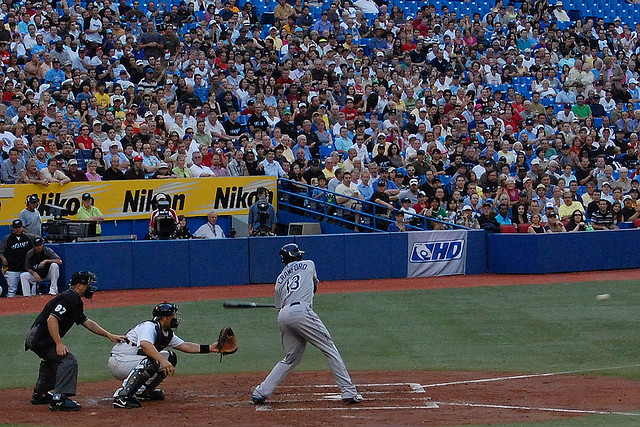}
    \caption{Image associated to the $\mathcal{R}$ and $c$ set defined in \ref{sec:examples}. (Taken from MS-COCO training dataset).}
    \label{fig:baseball}
\end{figure}

Finally, we computed the CLIPScore and CLIPScore$^{\text{ref}}$ for that set of reference and candidate sentences and its corresponding image (Figure \ref{fig:baseball}) with our own implementation of the metric.

\begin{equation*}
    \begin{aligned}
     \text{CLIPScore}(c,\mathcal{R})&= 0.7762\\
     \text{CLIPScore}^{\text{ref}}(c,\mathcal{R})&= 0.8387
    \end{aligned}
\end{equation*}

The value range from CLIPScore is expected to be from 0 (worst) to 1 (best). SoTA models achieve values around $0.76$ and $0.80$, respectively \cite{Stefanini2021}.

With this example, we show how different all the metrics are, reaching a very low score with ${BLEU}_4$ and SacreBLEU. On contrary, a very good score with SPICE, BERTScore, and the two metrics based on CLIP. We think most of us will consider the candidate caption to be adequate to the image, which is something that not all metrics do.

\subsection{Metrics put to the test}

With this variety of metrics, it is not easy to declare when a model is better than others, primarily because each metric measures different things, and it is essential to know how much we should trust the scores they produce. Measures of correlation between human judgment and metrics are presented in this section.

Maybe the most used but more evaluated and analyzed metric is BLEU since it was one of the first metrics used for the IC problem. A specific evaluation of it could be found in the work of Reiter in \cite{reiter2018structured}, where an exhaustive review of BLEU is presented, and as the main discussion, Reiter said that BLEU should not be the primary evaluation metric when it is necessary to present evidence to the research community.

Also, a comparison between metrics was made by Elliot et al. in \cite{elliott2014comparing} where Spearman's correlation of BLEU, ROUGE, TER, and METEOR with human judgment was measured. They used the Flickr8k dataset and found that BLEU and ROUGE had little correlation, TER even had a negative correlation, and only METEOR had a strong correlation.

In \cite{tanti2018pre}, Tanti et al. noted that the caption generation is slow as the model needs to generate one word at a time, and this process is multiplied if beam search is used. So, to evaluate a model, the captions need to have been generated; they called this form of evaluation post-gen. 

They presented a pre-gen way to evaluate the model as the caption is being generated and to consider the probability distribution of the entire vocabulary at each time step. They showed how this pre-gen metric correlates well with traditional post-gen ones to be used as a cheaper alternative. The drawback of this pre-gen metric is that it needs the entire model to evaluate, compared to the post-gen, where it can be done with just the pre-computed captions.

Following the work of \cite{elliott2014comparing}, Kilickaya et al. presented in \cite{kilickaya2016re} an extended correlation-based evaluation of the metrics BLEU, METEOR, ROUGE, CIDEr, SPICE, and Word Mover's Distance (WMD) using the Flickr8k and composite datasets. 

They reported the Pearson, Spearman, and Kendall correlation of all the metrics and William's test to verify that the metrics are statistically different. They conclude that the metrics can be combined to complement each other. They also conducted experiments on the Flickr30k dataset, replacing some elements of the captions, like the person involved or the place, to try to fool the metrics. One of their findings is that WMD had higher accuracy in detecting this variation.

The discrepancy of the automatic metrics with human ranking is a well-known problem in the NLP research community, such as in the Natural Language Generation (NLG). In \cite{novikova-etal-2017-need}, the authors showed how the metrics are poorly correlated with humans, affected by the dataset. They conclude that the metrics can distinguish between harmful and excellent sentences but not between medium and good quality.

Another recent study in NLG evaluation from Shimorina \cite{shimorina2018human} shows that the correlation of the metrics with humans depends on whether it is measured on model or sentence level; that is, when automatic metrics are used, they have a reasonable correlation with humans when the scores of the whole sentences are averaged.

Nevertheless, when we take individual sentences from different models, rank them, and score the models based on the ranks of their sentences, the result correlates less with human judgment. Their findings gave more evidence for the necessity of new and improved metrics.

During the MS-COCO Captions Challenge 2015, some image caption models produced results that outperformed the human-generated captions according to the metrics BLEU, METEOR, ROUGE, and CIDEr, but when humans acted as judges, none of the models were superior \cite{bernardi2016automatic}. Humans can not constantly evaluate the models because of their high cost, so it is essential to have reliable automatic metrics; the above indicates that the metrics should improve, at least for the image caption task.

The problems already mentioned with the metrics can worsen in datasets like Twitter, which has informal writing, and Ubuntu, which has technical terms.

In \cite{liu-etal-2016-evaluate}, the authors put BLEU, METEOR, ROUGE, and a metric based on Word2Vec to test Dialogue Response Generation Models on these datasets. Their findings were that the correlation on Twitter was low and even lower on Ubuntu. They conclude that metrics based on comparing $n$-grams need a large set of references to be meaningful, and those metrics based on embeddings have more potential to be relevant, but the full context of the text needs to be considered.

The search for new metrics had produced proposals like TIGEr \cite{jiang2019tiger}, a model based on SCAN \cite{Lee_2018_ECCV} and trained on MS-COCO that takes not only the referenced captions but also the image to rate the candidate description. 

TIGEr associates the words in the captions with regions of the image to determine their similarity. In the experiments, the authors showed stronger correlations with human judgment than BLEU, ROUGE, METEOR, CIDEr, and SPICE on the Flickr8k and composite datasets.

Recently, Stefanini et al. in \cite{9706348} compiled the results of some of the best current models. They applied the traditional metrics plus the TIGEr, BERTScore, and CLIPScore in addition to some diversity indicators. They measured the model-level correlation of the metrics with CIDEr and found a linear correlation. They conclude that CIDEr is a good indicator of the model's performance compared to other metrics.

This work tries to answer \emph{How good are the metrics? How can we evaluate them?}, and the designed methodology we propose is presented in the following section.


\section{Metrics for discrimination: our proposed methodology}\label{sec:ranking}

In this section, we will discuss how the most recurred metrics evaluate the quality of a candidate caption. As we already stated, the main goal of a metric is to assess how well a {\it translation} between the image domain and the natural language domain is done. The standard way is to compare how a human-made annotation (or several annotations) concurs with the one predicted by the IC model.

The automatic metrics presented previously (in Section \ref{sec:metrics}) will now be measured by how well they perform on discriminating {\it wrong} captions against {\it good} captions. 

The evaluation of the metric will be stepped into 4 degrees of performance, from worse to better, and we will know for sure their quality; this will be explained in the following subsection. 

The performance of arranging the captions from worse to better will be taken as a measure of how well the metric gives meaningful insight into the proposed captions.

The methodology proposed is different from that commonly applied where the metric, or a set of its meta-parameters, is {\it tuned} to correlate with human judgment. This implies that we need human judges to evaluate each of the captions, which is costly in both time, and money. 
In our methodology, we synthetically corrupt the captions in different degrees. The main advantage is that, unlike most of the metric evaluations (\cite{elliott2014comparing, novikova-etal-2017-need,shimorina2018human, bernardi2016automatic}), we dispense with human judgment, which is the more costly process.

We based our methodology on the 2017 MS-COCO database \cite{lin2014microsoft}. 
The MS-COCO dataset is the most used; created in 2014, it contained 83K images for training and 41K for validation, each with five captions per image. After the community's feedback, the creators decided to change the distribution to 118k images for training and 5K for validation in 2017. The test set consists of 41k images, but their 40 captions per image are not public, and all the evaluations must be done through the official server access\footnote{\url{https://cocodataset.org/\#captions-eval}}. 

We took the validation set, which includes 5,000 images, with five humanly annotated captions (references) for each image. We then separated one of the references from the rest and constructed one set of 5K human-generated captions, and with the remaining captions, we formed a set of 4 annotations for each of the 5K images, considered our ground truth. 

We will consider the set of captions taken apart from the set as the {\it best} captions and call it \emph{Human} captions. It will be compared against other {\it generated} captions, as we will explain hereunder.

To generate slightly worse captions, we went with two different approaches. For the first one, we generated a dictionary of all the unique words repeated at least four times in the references. It is a total of 1039 unique words, which will be our bag of words. We then took 25 percent random words from the \emph{Human} captions, took them apart (the {\it best} captions), and substituted them with a random word from the bag-of-words. 

This will be our second best (called \emph{Replace 25\%})  set of captions. We did the same for 50\% random words to produce the set \emph{Replace 50\%}, which will be the third-best set of captions. For the {\it worse} captions, instead of randomly taking words from the dictionary to generate a nonsense sentence, we decided to randomly take another human-generated caption that does not correspond to the image to be captioned. 

This set of captions is called \emph{Random}. For instance, if the image had a humanly annotated caption of {\it a girl blows candles from a cake}, we will assign a random caption such as: {\it a zebra is grazing on a green hill}. It is worth mentioning that even though the assigned caption does not correspond to the image, it is still a sentence that has intrinsic meaning and a human-generated. 

\begin{figure}
    \centering
    \includegraphics[width=0.75\textwidth]{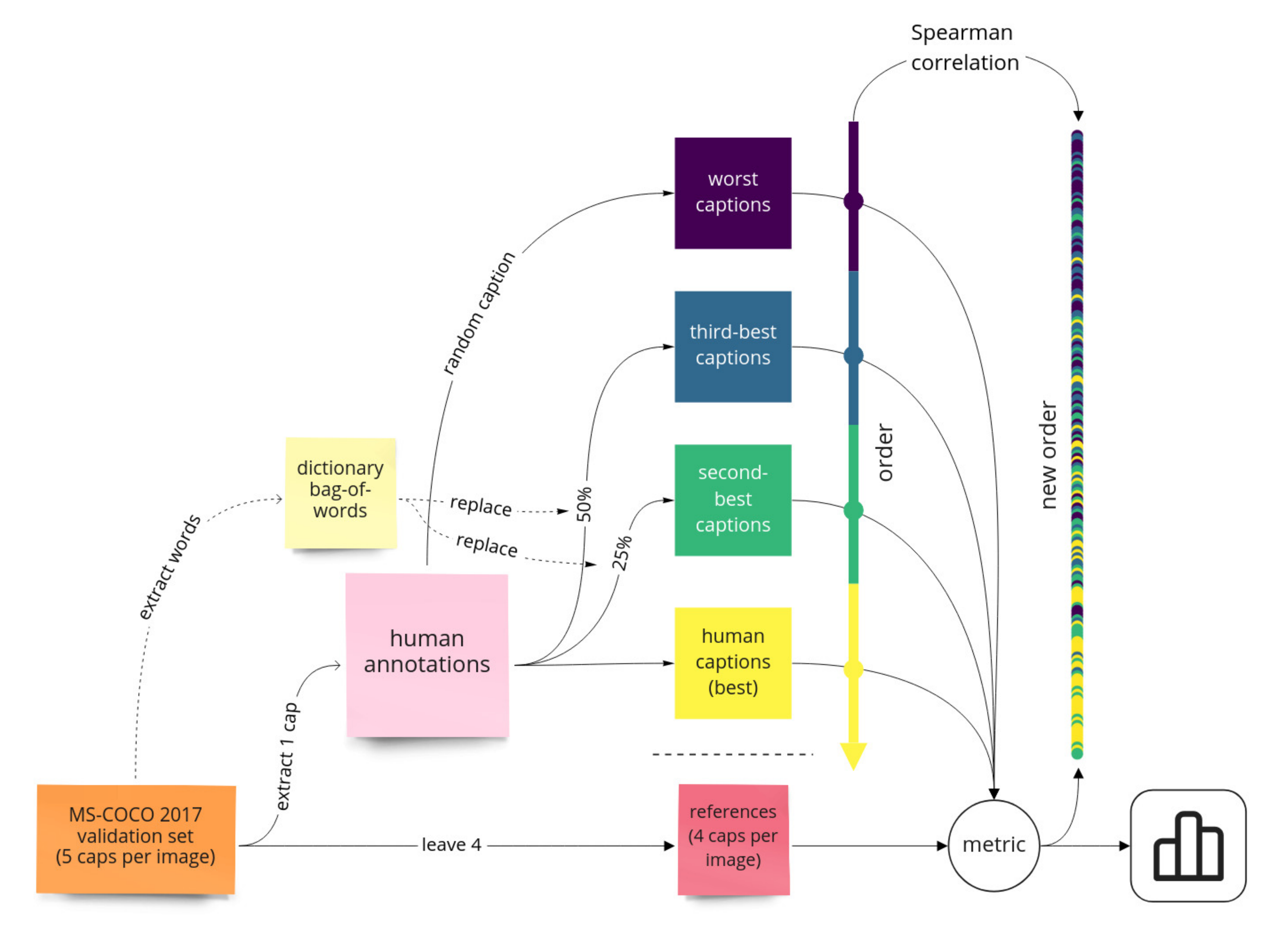}
    \caption{This is the general scheme in which we take the 5 references of the MS-COCO dataset and subdivide it into human annotations ({\it best} captions), and a subset of human annotations. We then subsequently distort the latter to generate artificial worse captions. In this scenario we replace words with another random word taken from a dictionary of all the annotations.}
    \label{fig:order}
\end{figure}

The methodology is illustrated in Figure \ref{fig:order}, where one can easily observe how the four sets of captions were generated and evaluated.

The expected behavior for all the metrics is that they could be able to both discriminate between the artificially generated caption and also order them from worse to better. We present here the results of our findings by plotting the order in which the metrics assigned a score to each of the captions generated by each different strategies, but also assigning them a color depending on if they belong to the \emph{Random, Replace 25\%, Replace 50\%}, or \emph{Human} captions, as indicated in the plots. The plots are histograms that depict the distribution of the captions for each configuration. We also calculated the Spearman correlation coefficient on the order given to the sets of captions, to get a single number that reports the overall performance of the metric. As it is known the Spearman correlation coefficient will tell us how monotonic the new order is given to the samples after ordering them by the metric score. Ideally, the order should remain the same, therefore a value of $+1$ would be expected. The equation that calculates the Spearman correlation $\rho_s$ is

\begin{equation}
\rho_{s}=1-{\frac {6\sum d_{i}^{2}}{n(n^{2}-1)}\,\,},
\end{equation}

\noindent where $d_i$ is the difference between the two ranks of each sample, and $n$ is the number of samples.

\begin{figure}
  \centering
  \subfloat[BLEU-1 $(\rho_s= 0.640800)$]{\includegraphics[width=0.37\textwidth]{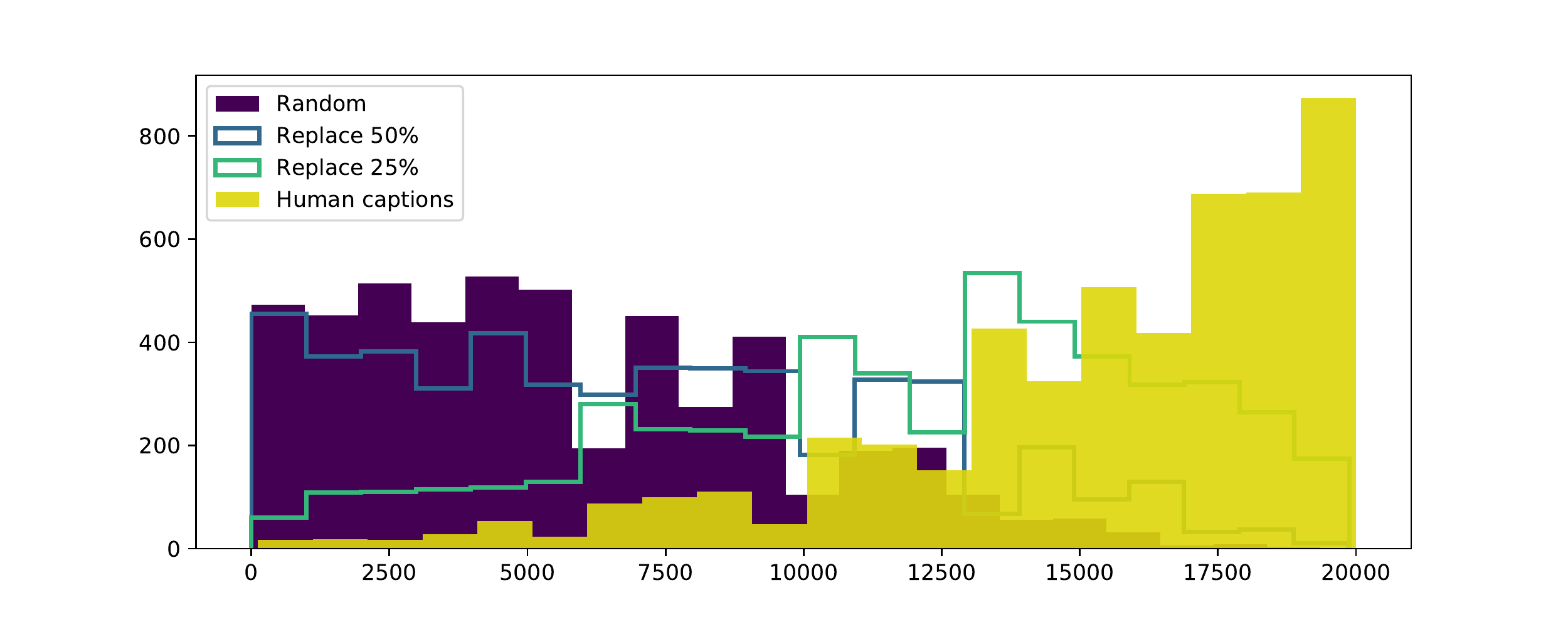}}
  \subfloat[BLEU-2 $(\rho_s= 0.555879)$]{\includegraphics[width=0.37\textwidth]{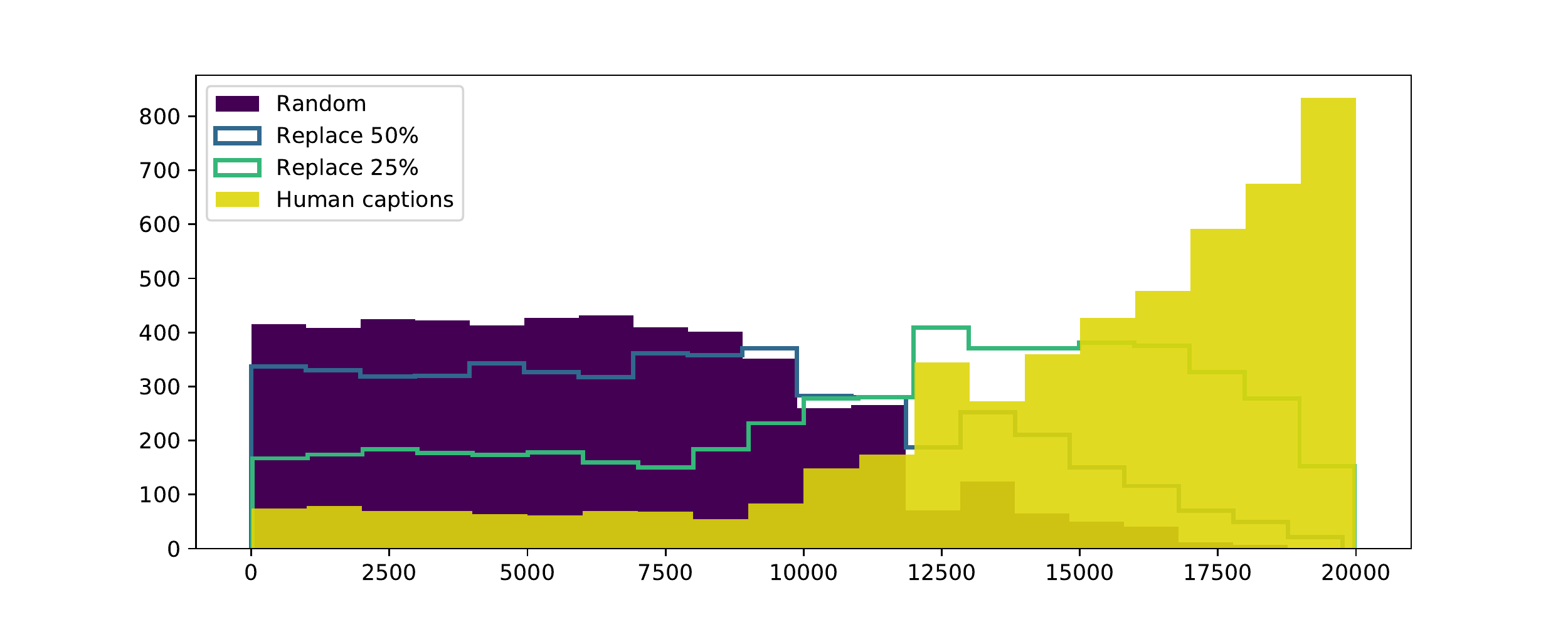}}\hfill
  \subfloat[BLEU-3 $(\rho_s= 0.330240)$]{\includegraphics[width=0.37\textwidth]{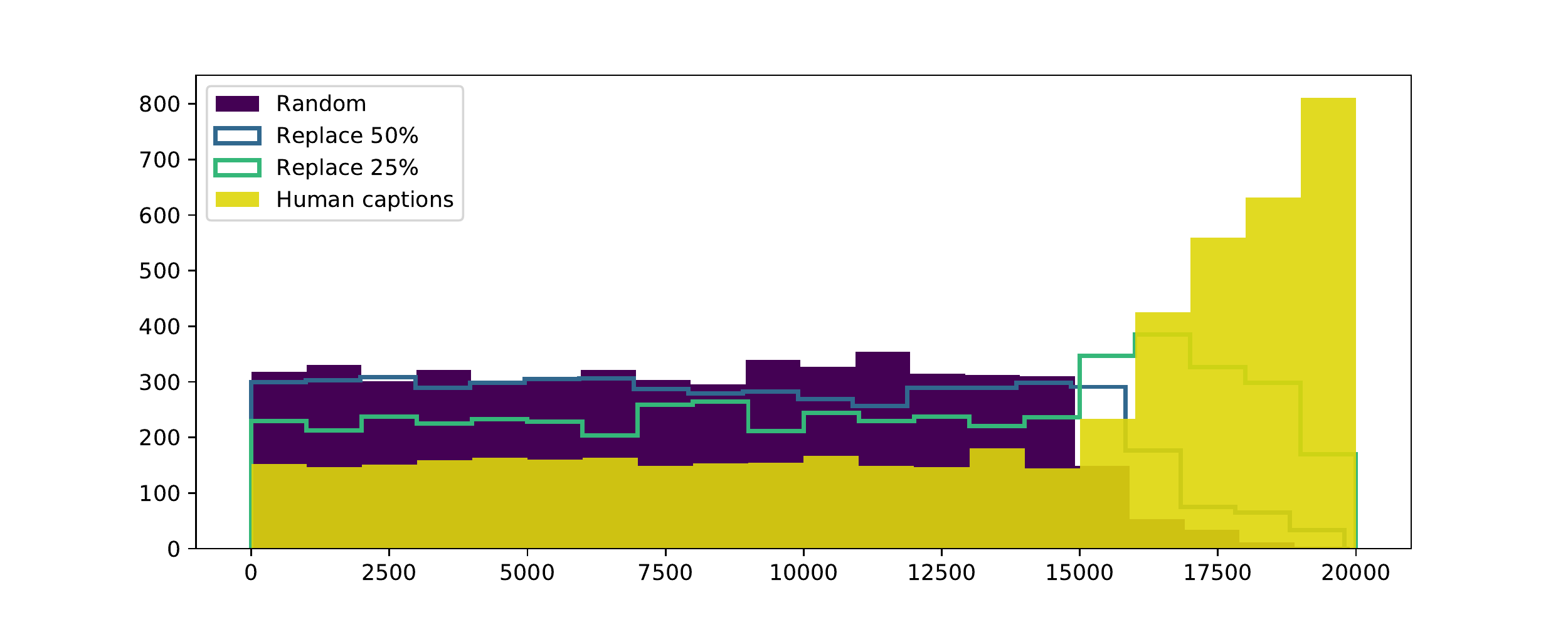}}
  \subfloat[BLEU-4 $(\rho_s= 0.128960)$]{\includegraphics[width=0.37\textwidth]{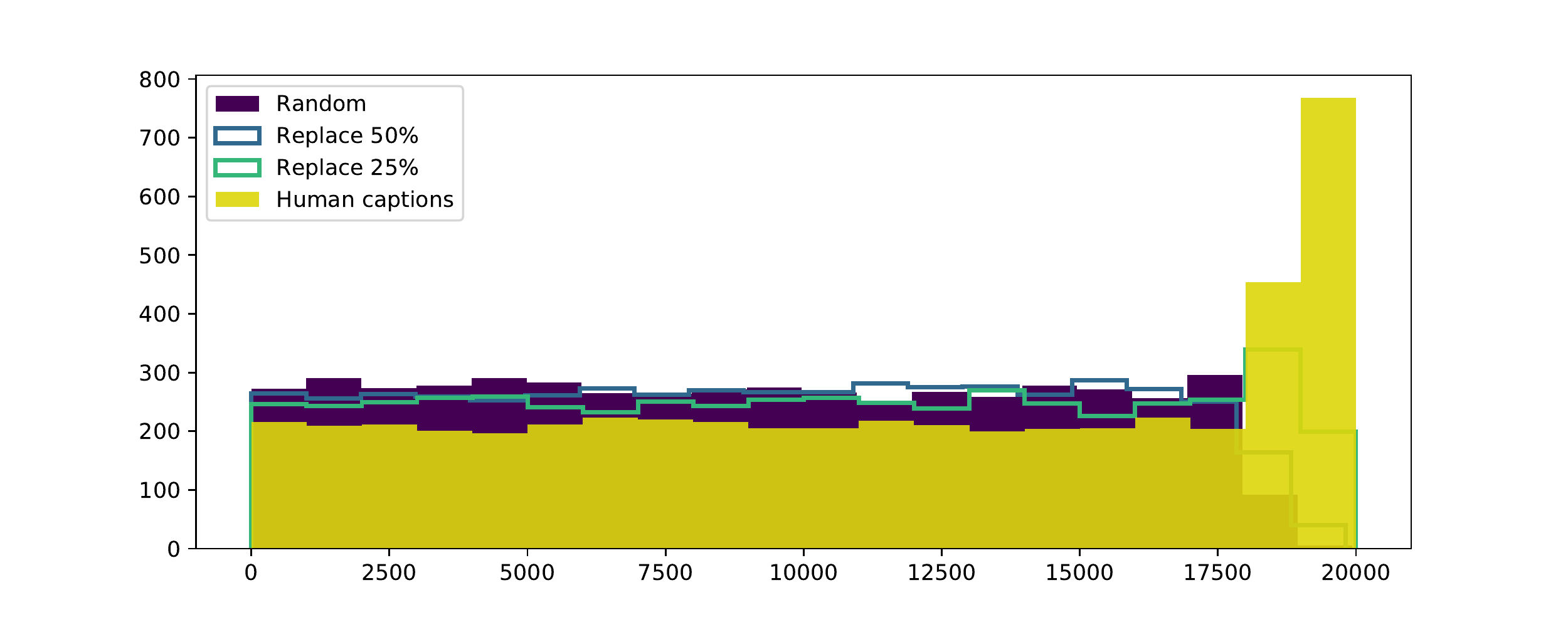}}\hfill
  \subfloat[SacreBLEU $(\rho_s= 0.560640)$]{\includegraphics[width=0.37\textwidth]{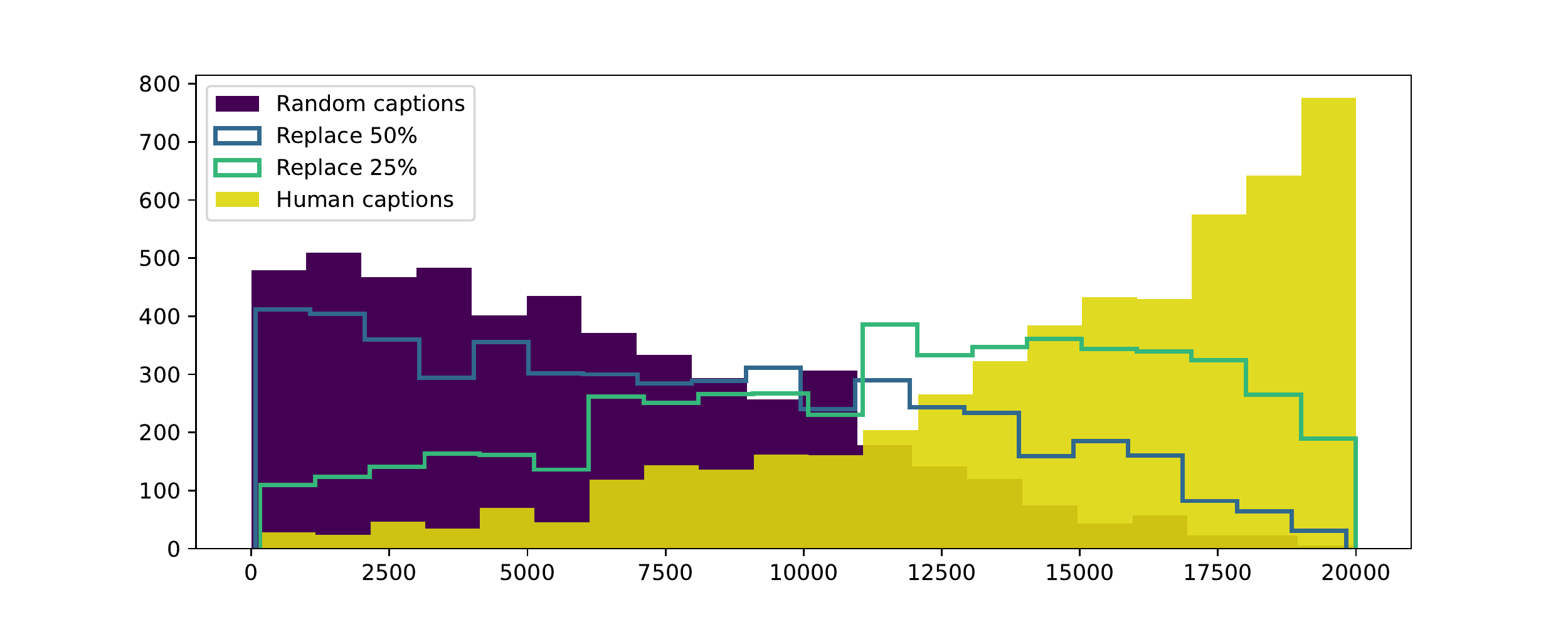}}
  \subfloat[METEOR $(\rho_s= 0.575520)$]{\includegraphics[width=0.37\textwidth]{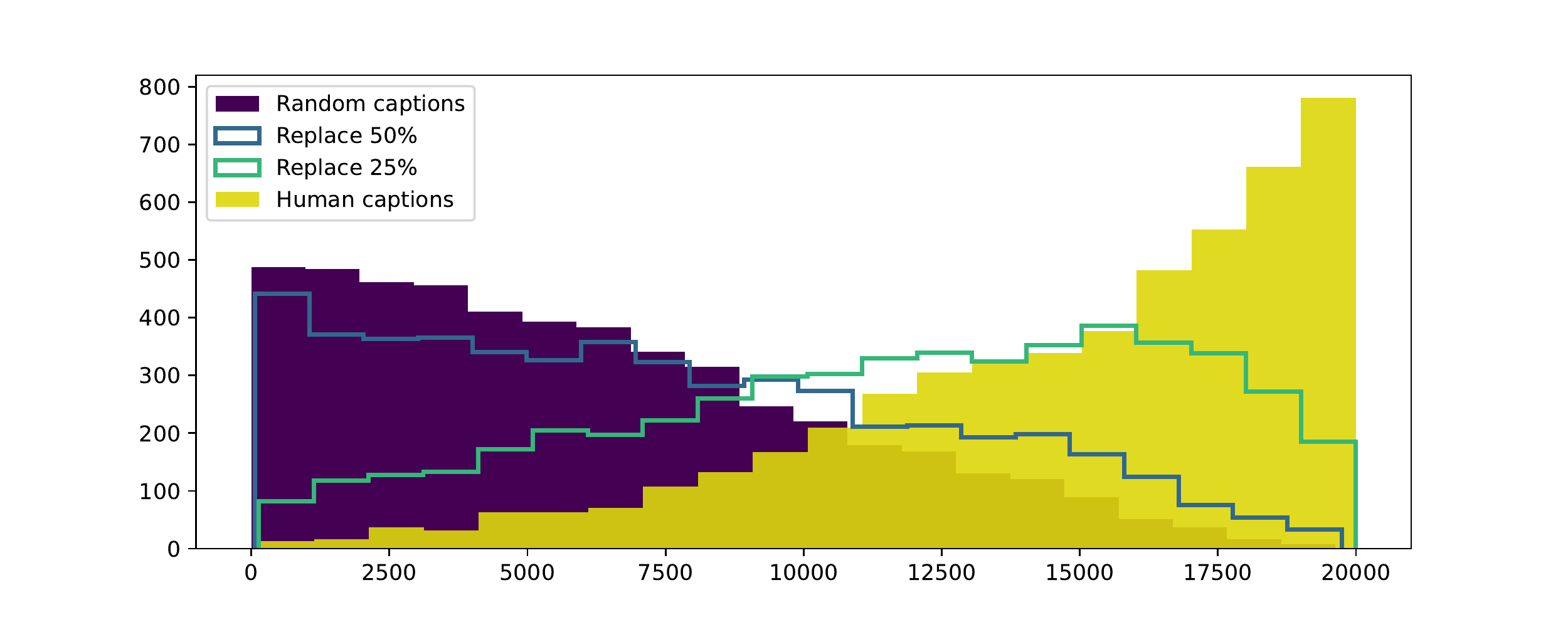}}\hfill
  \subfloat[SPICE $(\rho_s= 0.661920)$]{\includegraphics[width=0.37\textwidth]{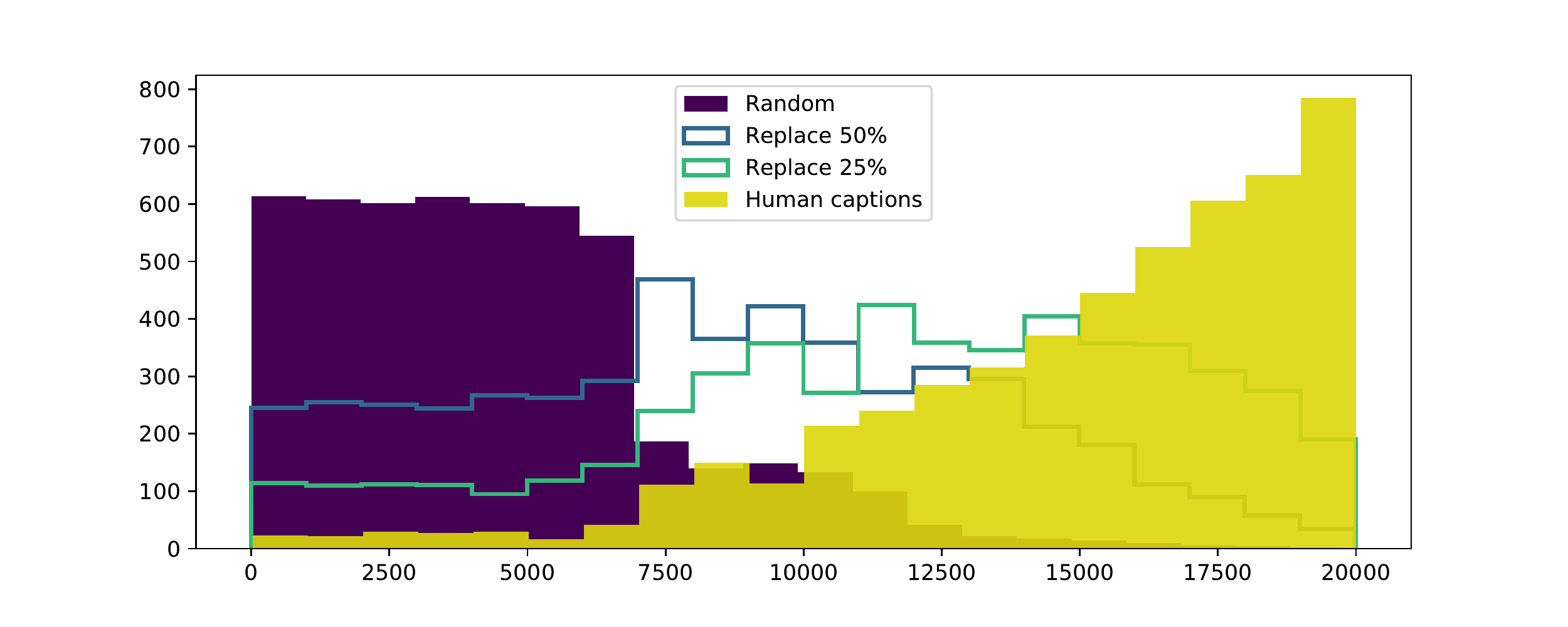}}
  \subfloat[CIDEr $(\rho_s= 0.708320)$]{\includegraphics[width=0.37\textwidth]{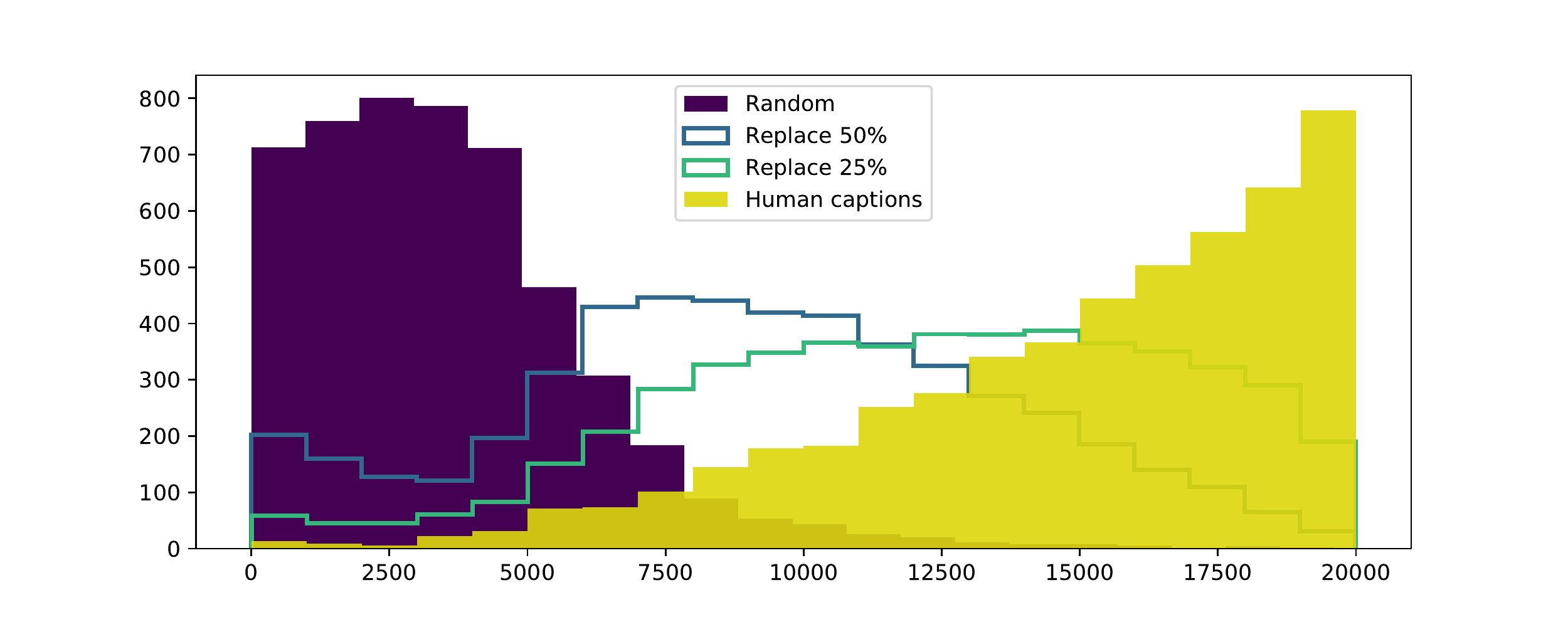}}\hfill
  \subfloat[ROUGE-L $(\rho_s= 0.570840)$]{\includegraphics[width=0.37\textwidth]{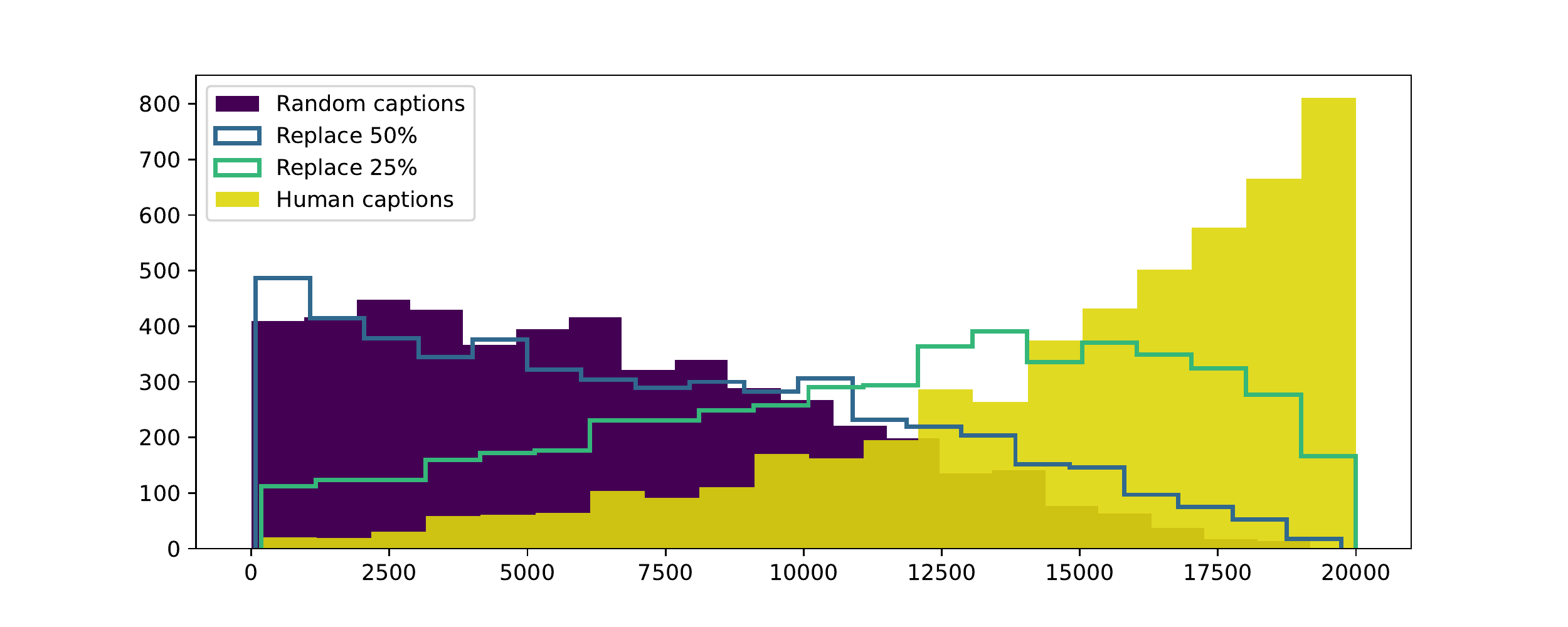}}
  \subfloat[BERTScore $(\rho_s= 0.556920)$]{\includegraphics[width=0.37\textwidth]{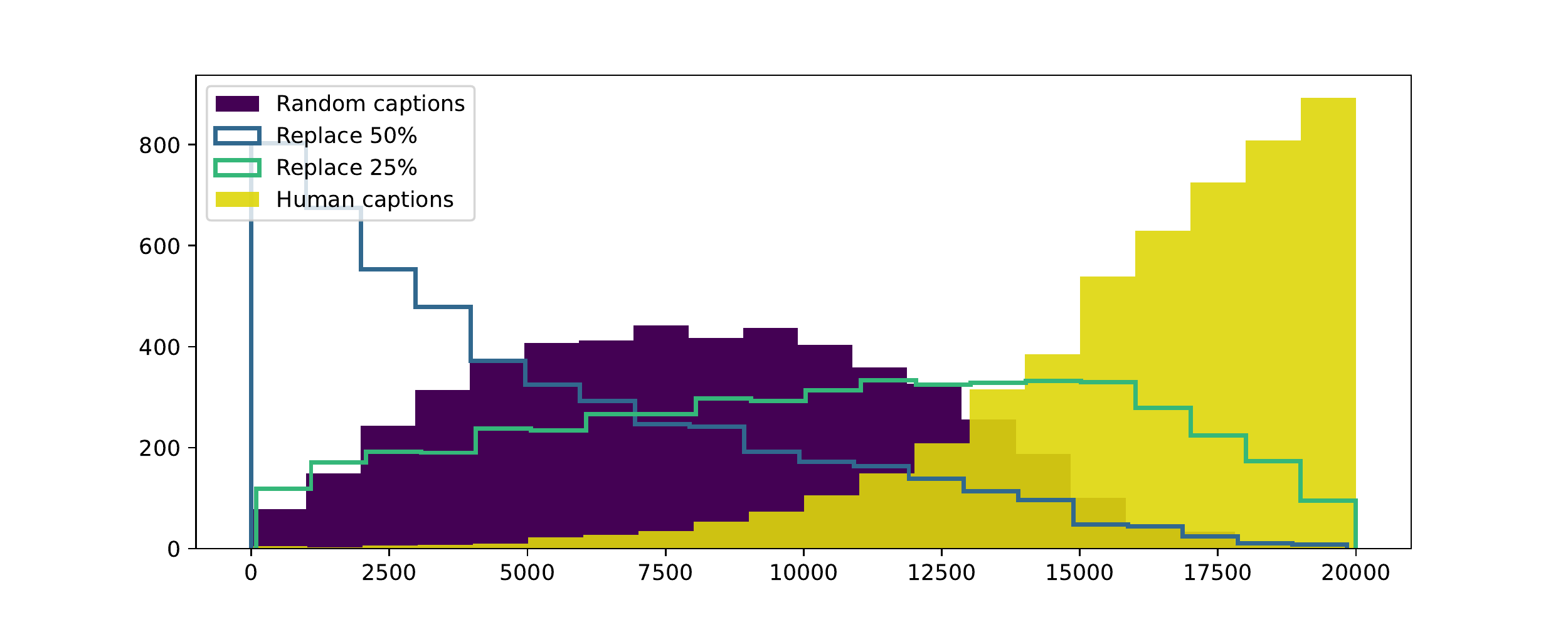}}\hfill
  \subfloat[CLIPScore $(\rho_s= 0.774080)$]{\includegraphics[width=0.37\textwidth]{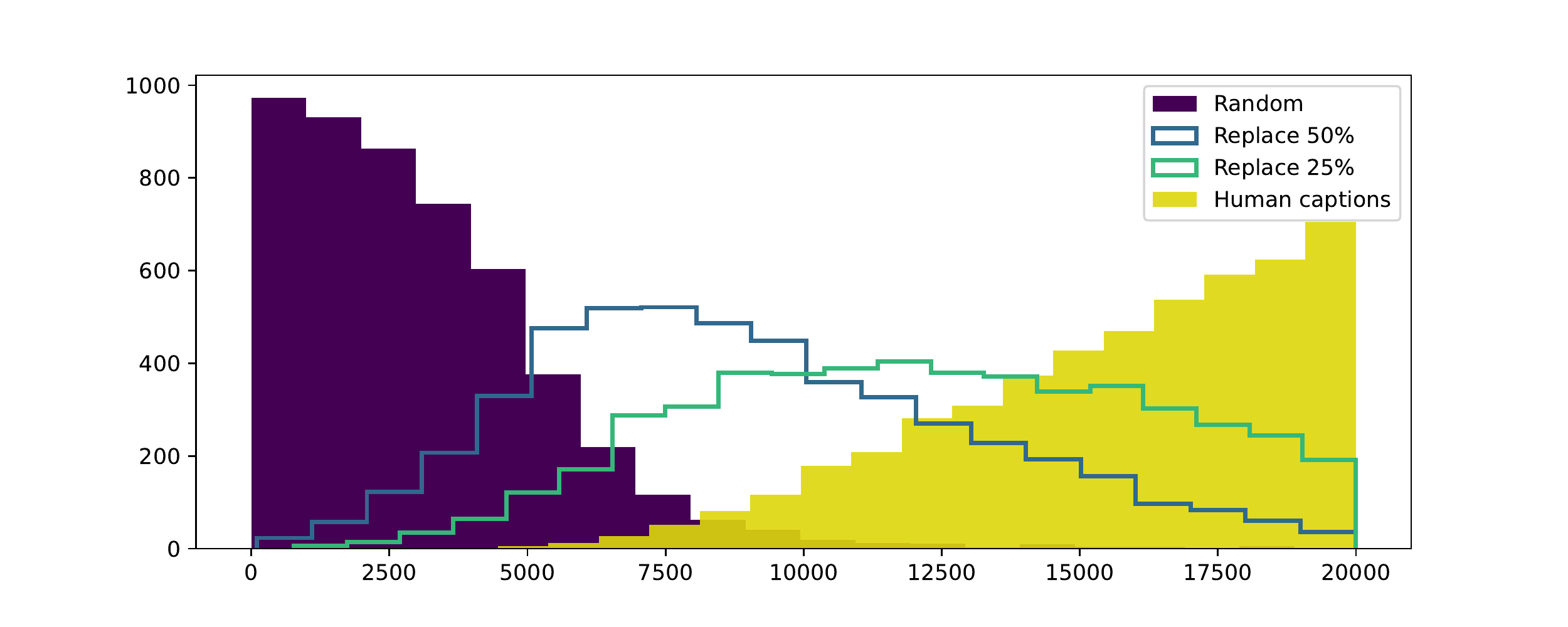}}
  \subfloat[CLIPScore$^{\text{ref}}$ $(\rho_s= 0.805640)$]{\includegraphics[width=0.37\textwidth]{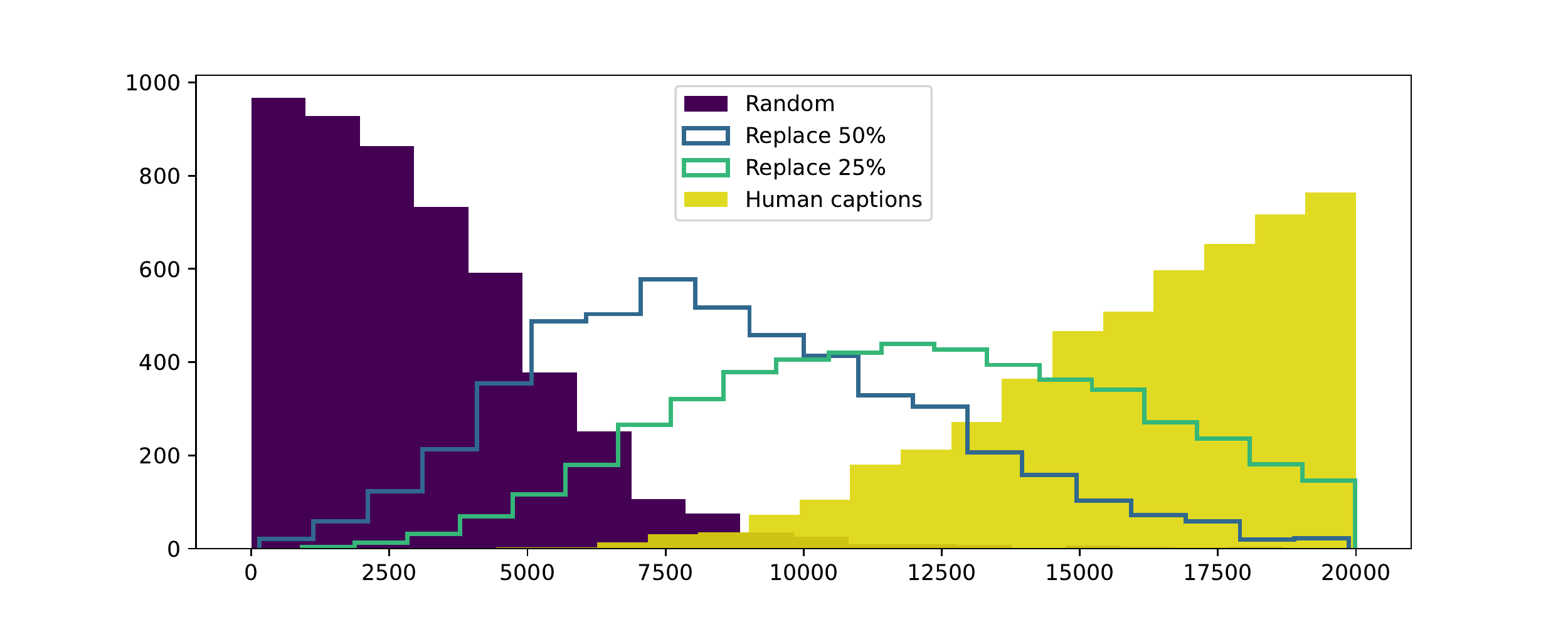}}
  \caption{Replacing methodology.\label{fig:replace}}
\end{figure}

The different order distributions are plotted in Figure~\ref{fig:replace}. We can see that the BLEU-1 does a better job separating the captions than the rest of the BLEU's. It is clear that the intersection of the histograms for the \emph{Random} and \emph{Human} captions increases from BLEU-1 to BLEU-4. This behavior can be expected because for larger $q$-grams, it is very likely for a replaced word to break the matches. SacreBLEU does a better job because it takes the longest matching $q$-gram, so it has more options to assign a score. METEOR and ROUGE-L behave similarly to SacreBLEU. With the SPICE metric we can see that the intersection of the Random and Human captions shrinks a little but there are a lot of captions for \emph{Replace $50\%$} and \emph{Replace $25\%$} that are scored very low. With CIDEr we can see that the \emph{waves} for \emph{Replace $50\%$} and \emph{Replace $25\%$} take more defined shapes and move to the right. The best match was obtained with the CLIP and CLIP$^{\text{ref}}$ scores where we can see that the intersection of the \emph{Random} and \emph{Human} captions is very small and we can distinguish a clear bell shape for the \emph{Replace} $50\%$ and \emph{Replace $25\%$} sets. Something different happens with the BERTScore, the \emph{Random} captions got better scores than the \emph{Replace $50\%$} set. It appears that BERTScore prefers well-written candidates, even though the prediction does not correspond to the references, and hence, to what is seen on the image. Our hypothesis is that BERTScore prefers grammar correctness over caption accuracy when it is presented to a candidate caption.

\begin{figure}
    \centering
    \includegraphics[width=0.75\textwidth]{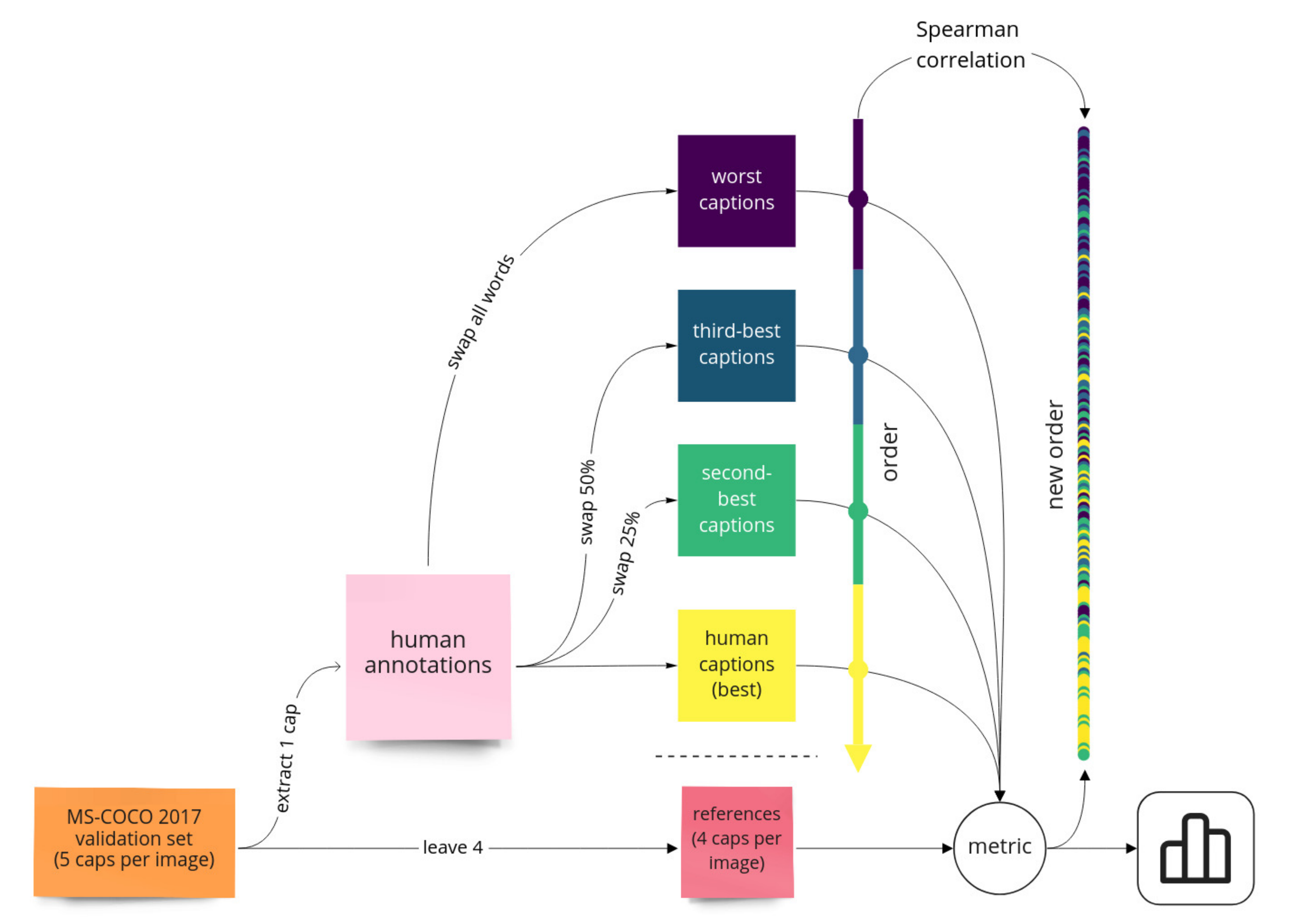}
    \caption{Following the ideas presented in Figure \ref{fig:order} we present another way to artificially generate worse captions to evaluate with the metrics discussed. We will now take the subset of human annotations and subsequently distort them to generate artificially worse captions. In this scenario we randomly swap the order of the words in the caption.}
    \label{fig:order2}
\end{figure}

To further explore this idea we decided to plot another scheme similar to the one presented in Figure \ref{fig:order}, but now, instead of replacing words within the caption with our bag-of-words, we decided to leave the same words from the original humanly-annotated caption and just swap the order of 25, 50, and 100 percent of the words (called \emph{Shuffle all, Shuffle 50\%}, and \emph{Shuffle 25\%}). If our hypothesis were correct, BERTScore would not be fooled by the worse captions in which the sentence lacks sense, but CLIPScore and CLIPScore$^{\text{ref}}$ may be, since the correct words, or elements of the caption are still present in the scrambled sentence. It is also our hypothesis that CLIP is a good metric for detecting features or objects on the scene, but it will not perform well for assessing the grammatical structure of the candidate sentence. The scheme of this second proposal is depicted in Figure \ref{fig:order2}.

\begin{figure}
  \centering
  \subfloat[BLEU-1 $(\rho_s= 0.027080)$]{\includegraphics[width=0.37\textwidth]{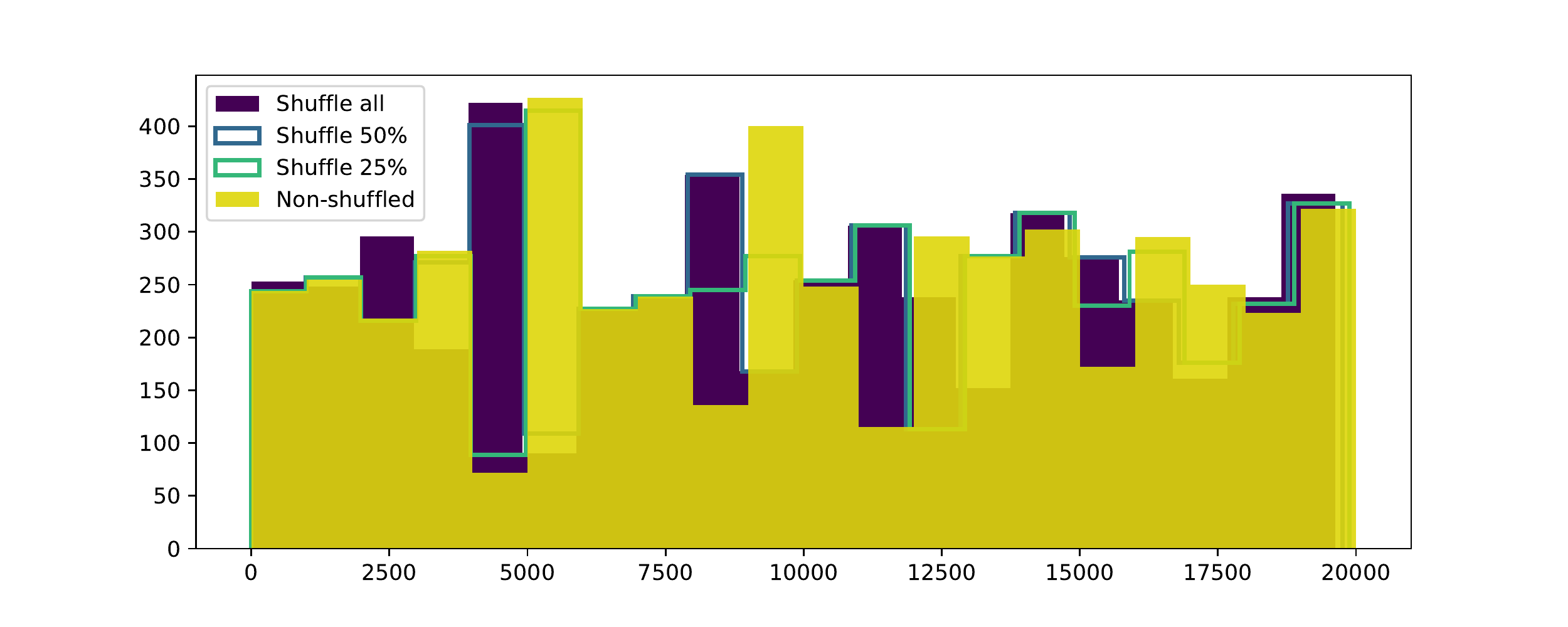}}
  \subfloat[BLEU-2 $(\rho_s= 0.380119)$]{\includegraphics[width=0.37\textwidth]{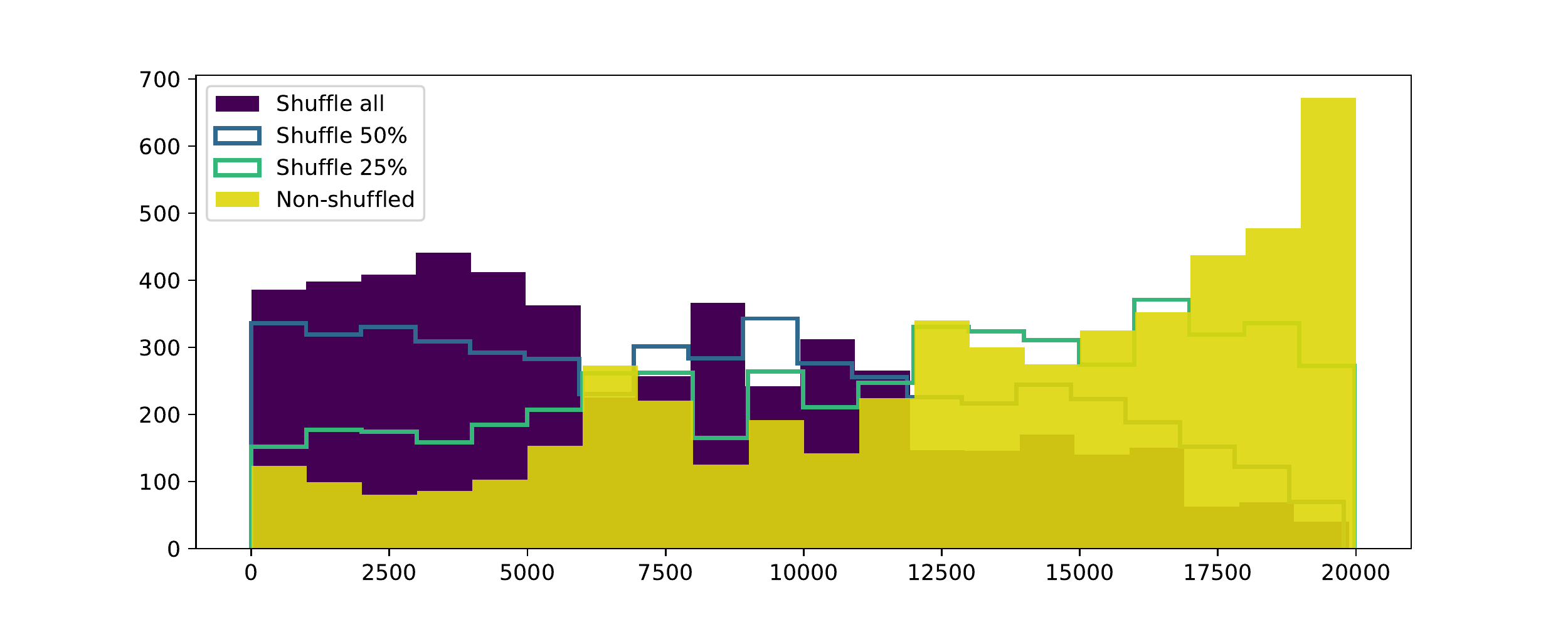}}\hfill
  \subfloat[BLEU-3 $(\rho_s= 0.338320)$]{\includegraphics[width=0.37\textwidth]{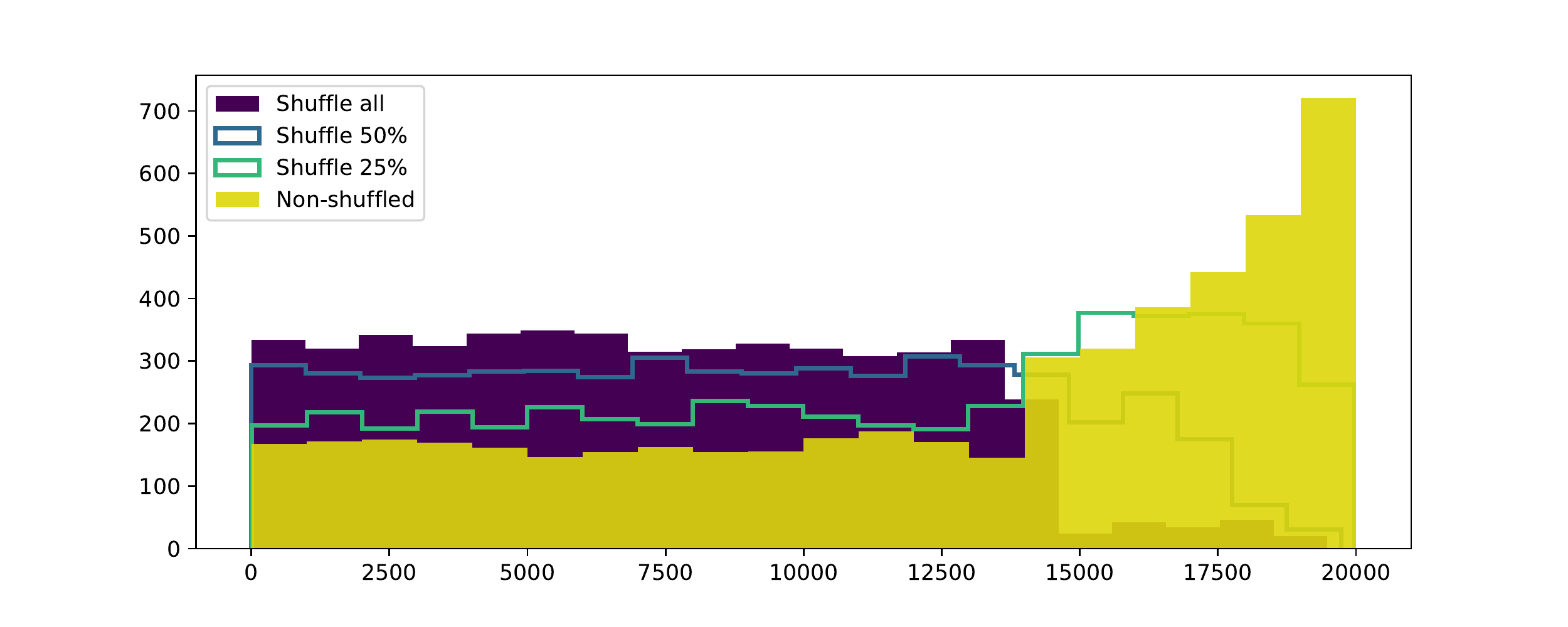}}
  \subfloat[BLEU-4 $(\rho_s= 0.134680)$]{\includegraphics[width=0.37\textwidth]{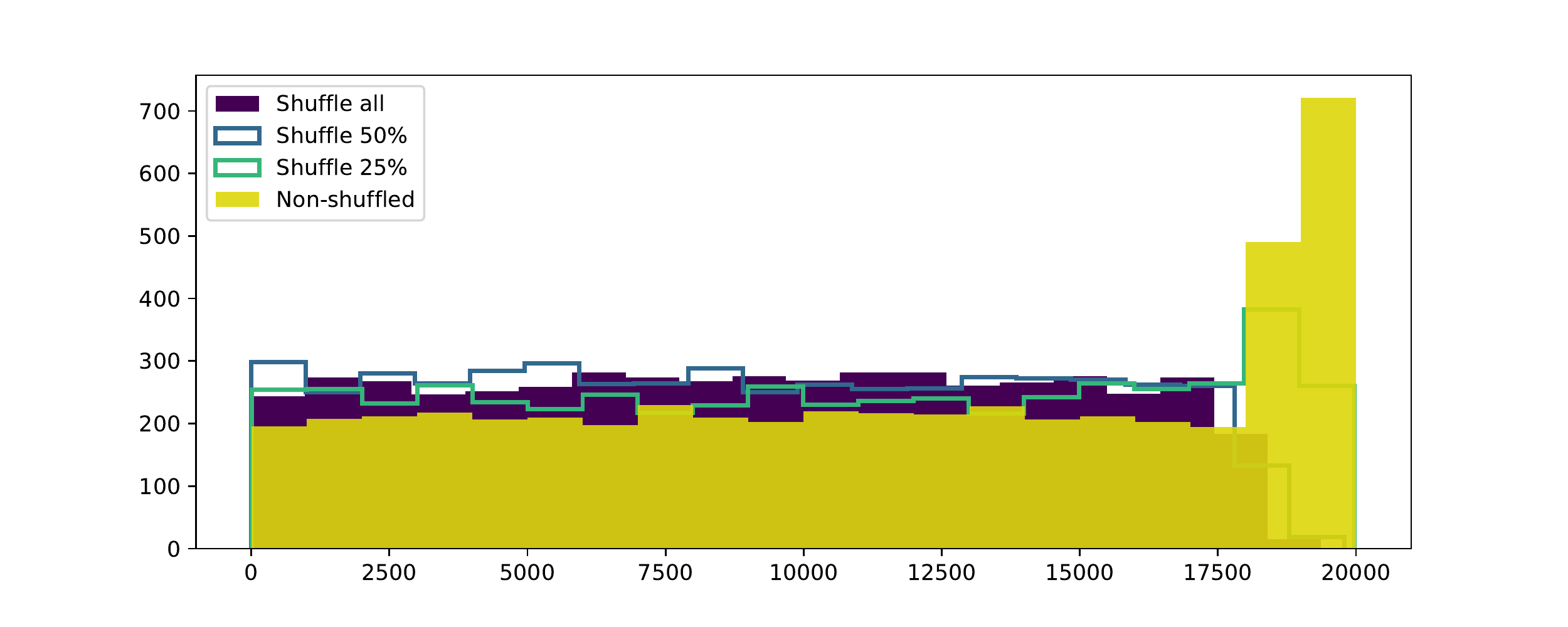}}\hfill
  \subfloat[SacreBLEU $(\rho_s= 0.474120)$]{\includegraphics[width=0.37\textwidth]{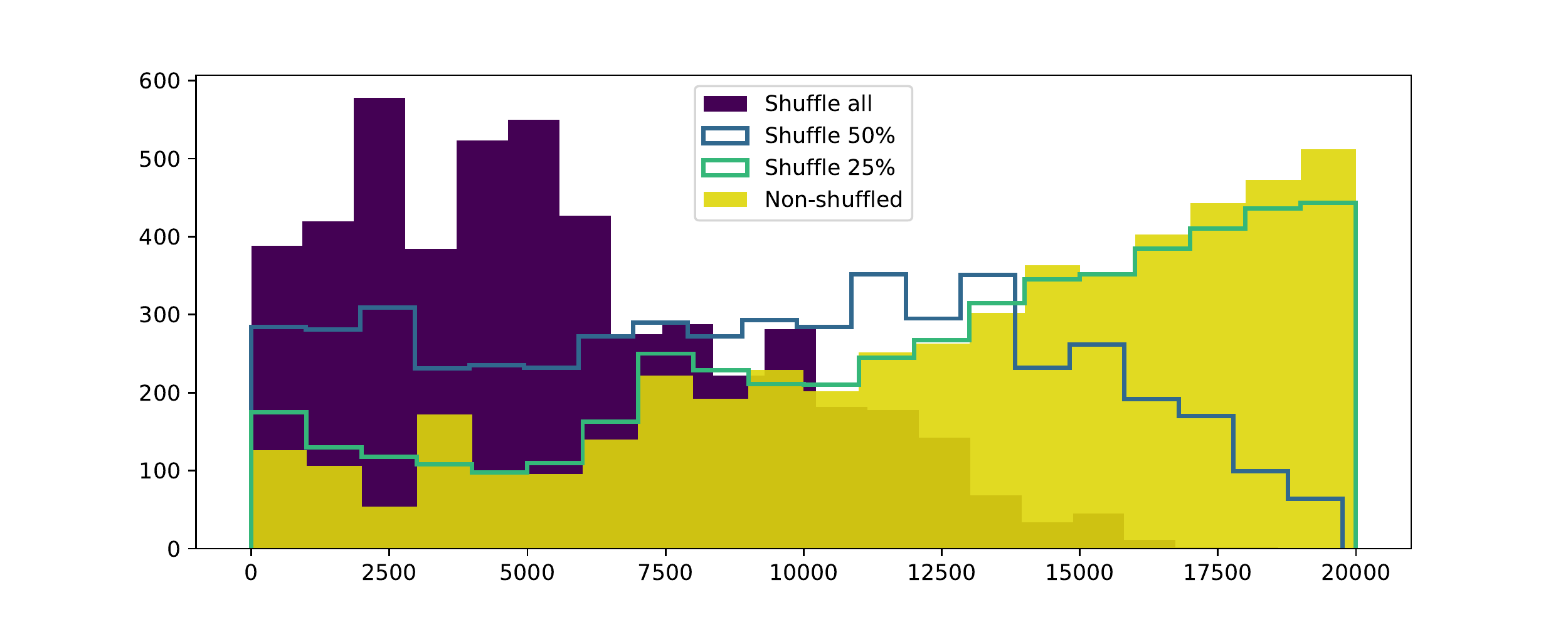}}
  \subfloat[METEOR $(\rho_s= 0.216000)$]{\includegraphics[width=0.37\textwidth]{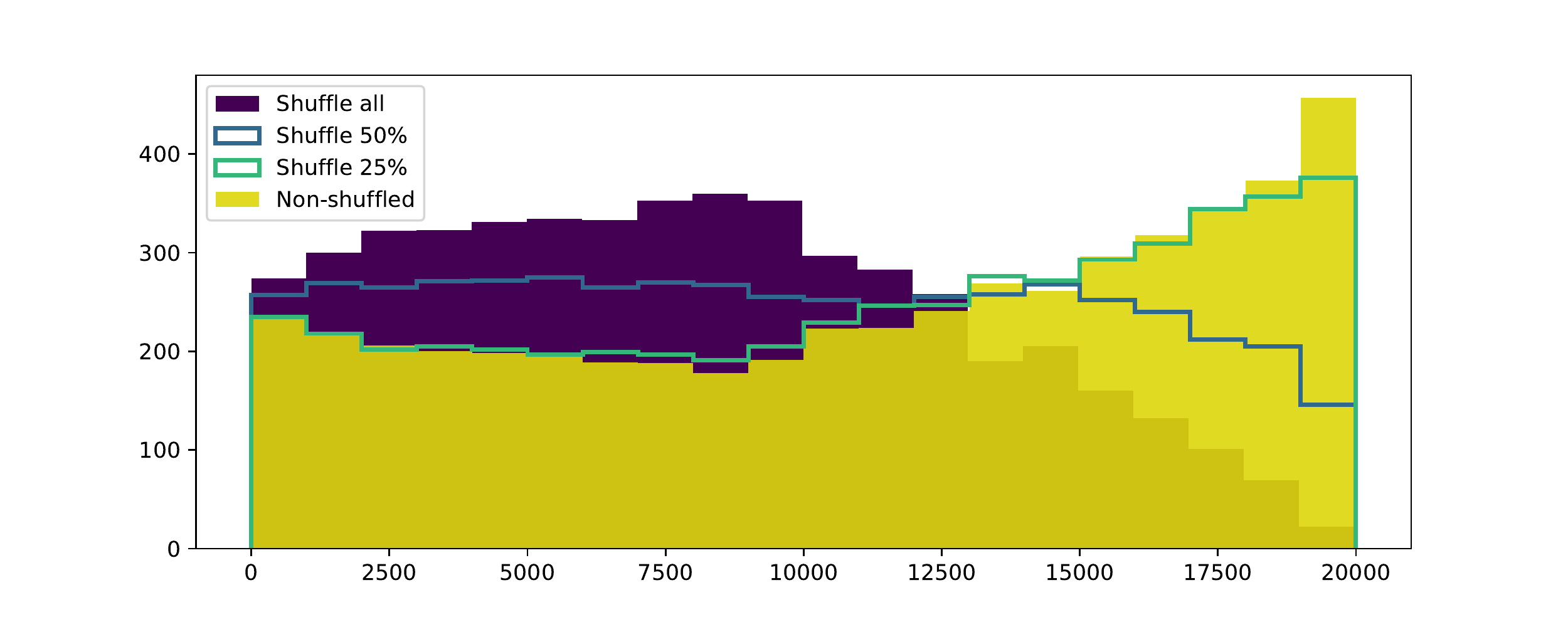}}\hfill
  \subfloat[SPICE $(\rho_s= 0.230680)$]{\includegraphics[width=0.37\textwidth]{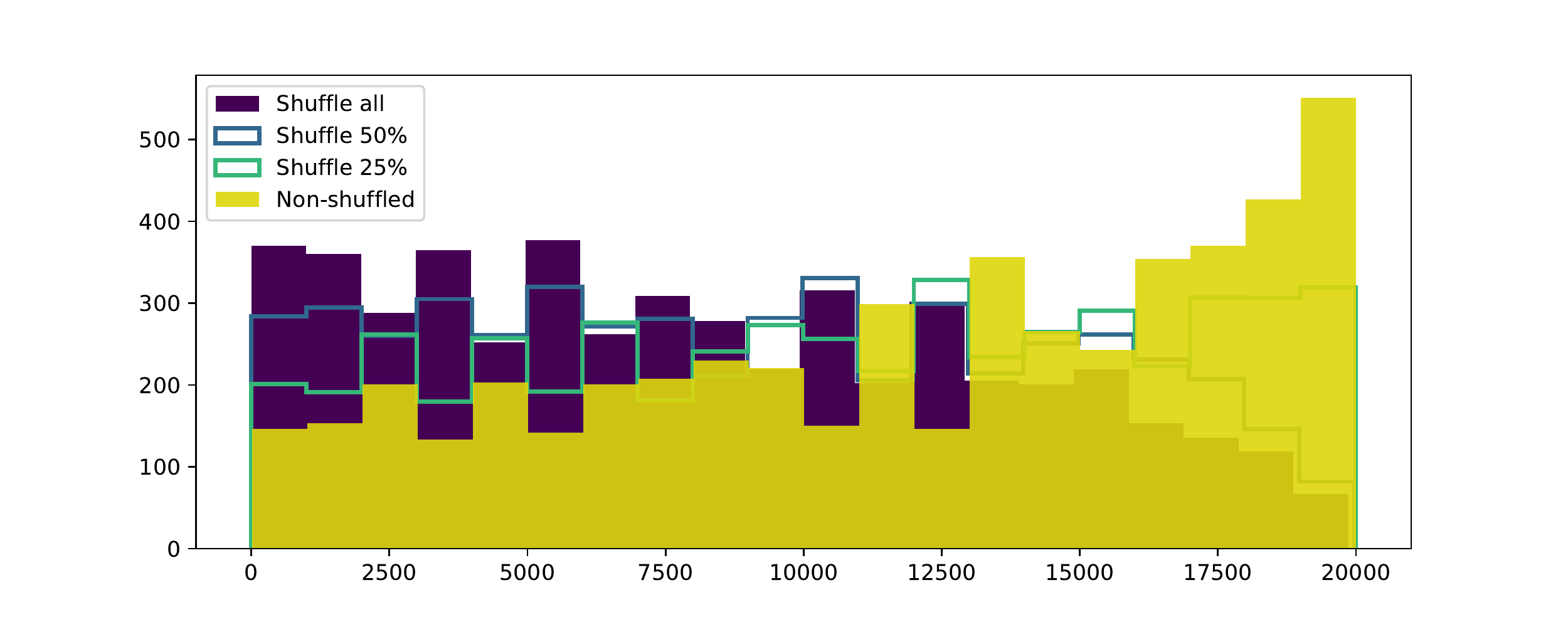}}
  \subfloat[CIDEr $(\rho_s= 0.123240)$]{\includegraphics[width=0.37\textwidth]{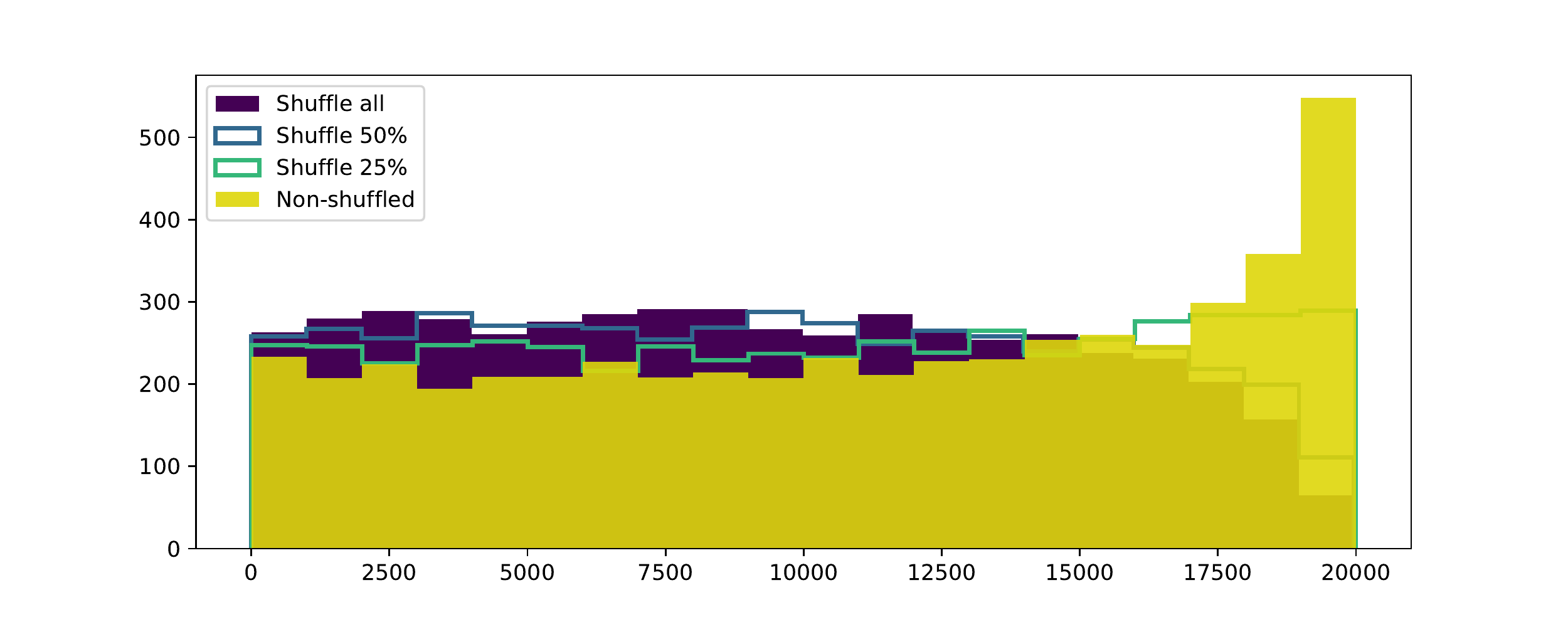}}\hfill
  \subfloat[ROUGE-L $(\rho_s= 0.171400)$]{\includegraphics[width=0.37\textwidth]{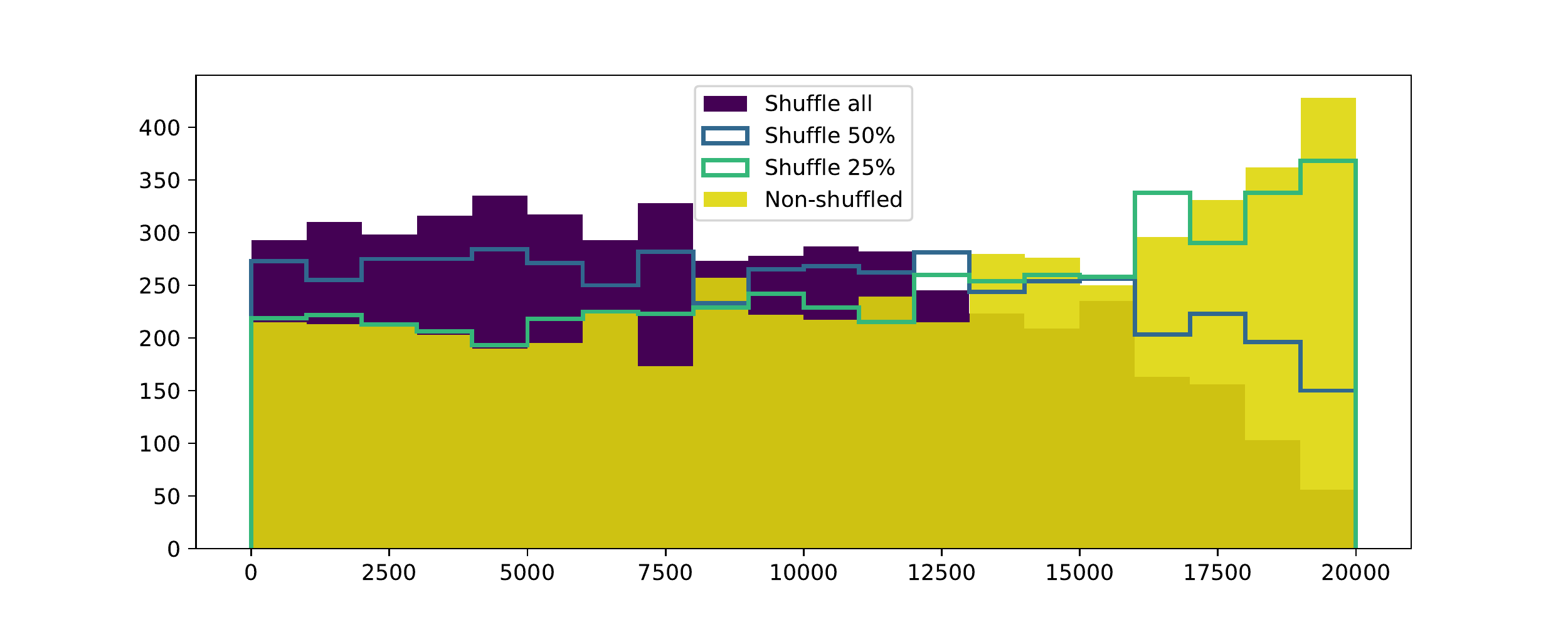}}
  \subfloat[BERTScore $(\rho_s= 0.709639)$]{\includegraphics[width=0.37\textwidth]{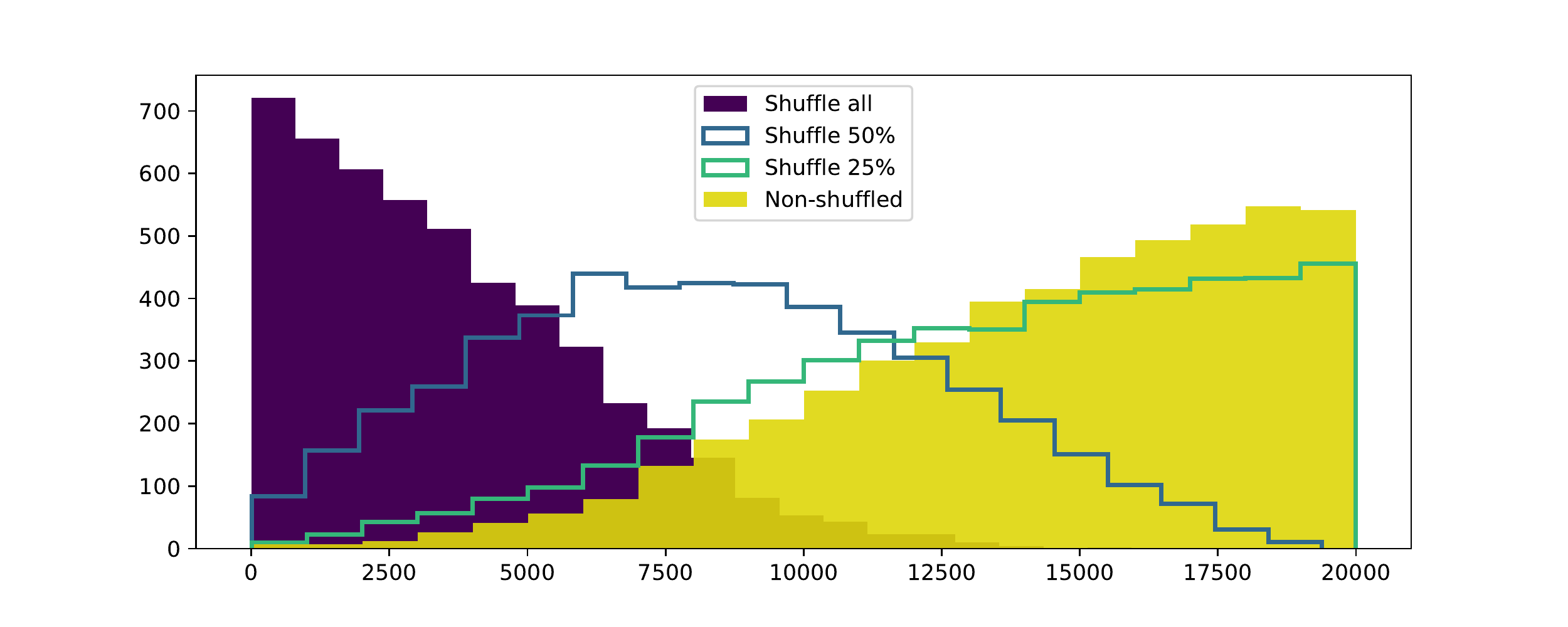}}\hfill
  \subfloat[CLIPScore $(\rho_s= 0.225760)$]{\includegraphics[width=0.37\textwidth]{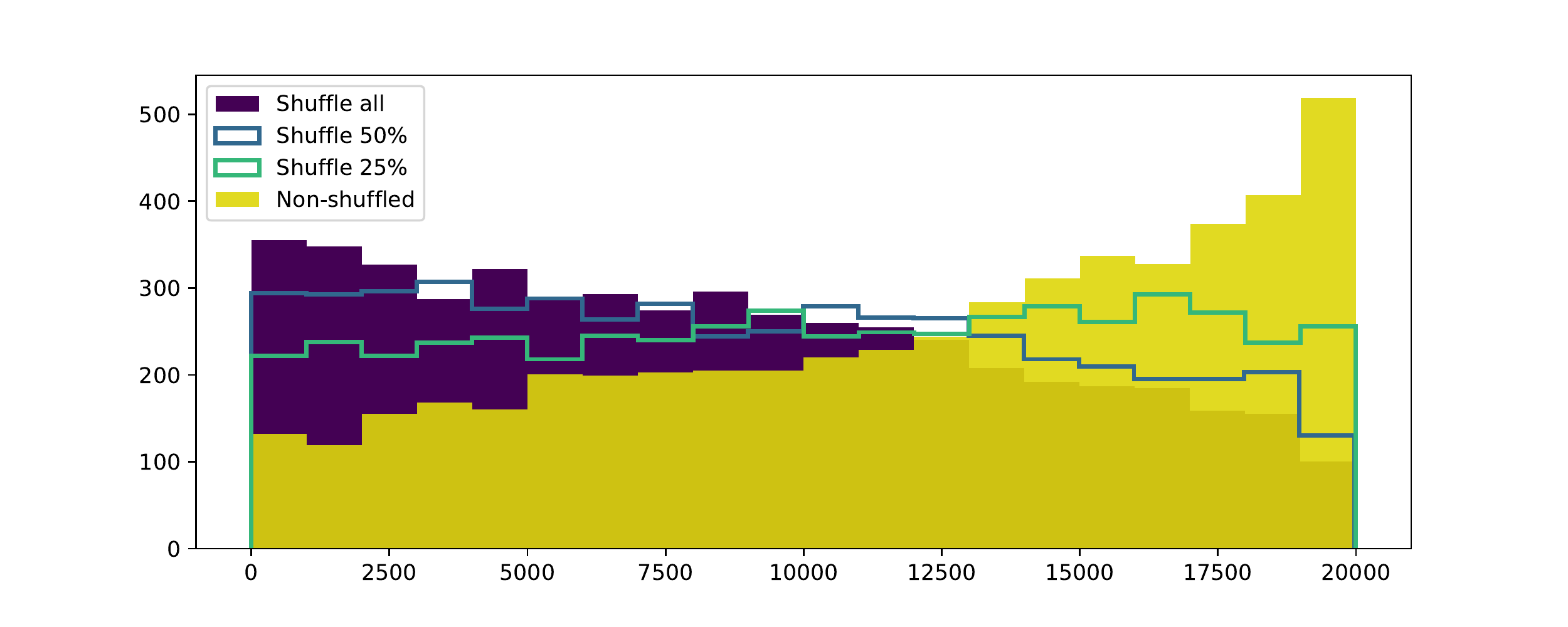}}
  \subfloat[CLIPScore$^{\text{ref}}$ $(\rho_s= 0.280159)$]{\includegraphics[width=0.37\textwidth]{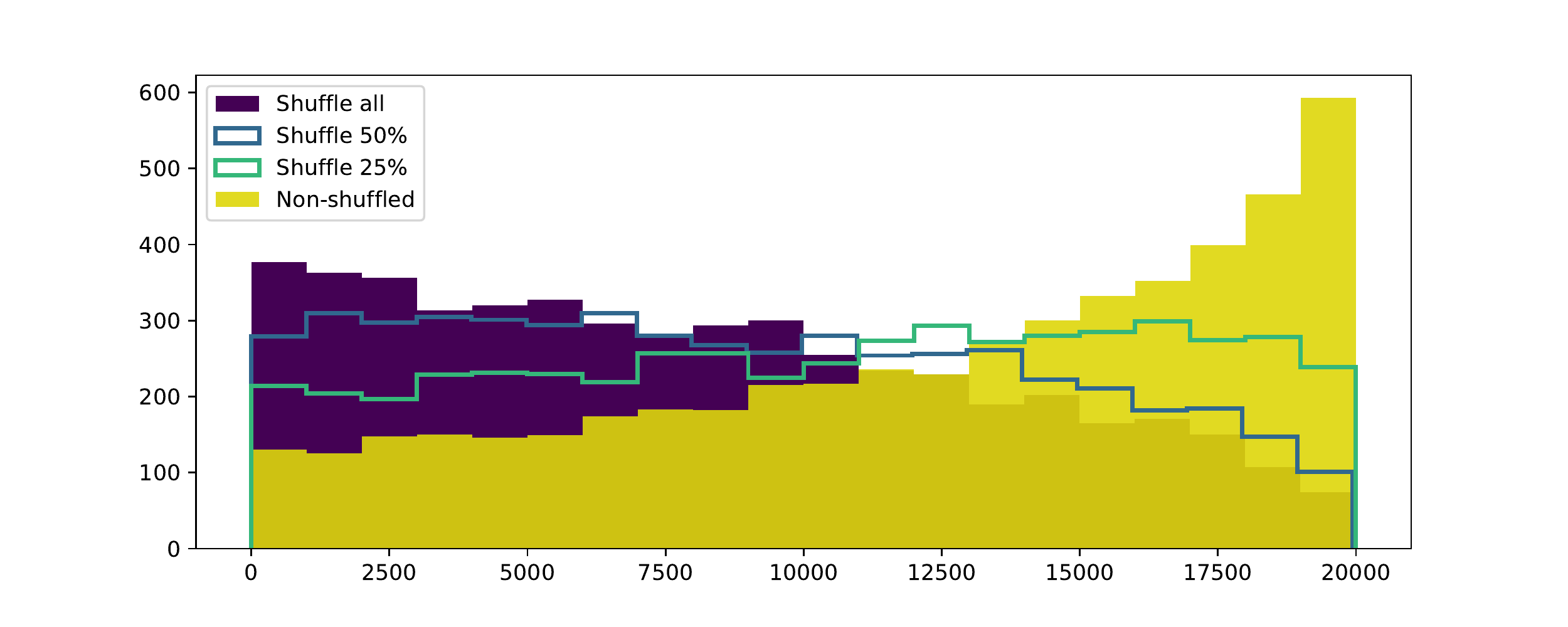}}
  \caption{Shuffling methodology.\label{fig:shuffle}}
\end{figure}

In Figure~\ref{fig:shuffle} we show the histograms for the order produced by each metric on the scrambled sentences. BLEU-1 can not distinguish between the different levels of shuffling, which is expected because it only cares for $q$-grams of length one, independent of the order. The best of this family is BLEU-2, a probabilistic argument can explain that there are more $q$-grams of length two than three or four. 

SacreBLEU got a better correlation, the reason is the same, it cares for the longest possible $q$-gram. METEOR, SPICE, CIDEr, and ROUGE-L all have a low correlation and are unable to discriminate between the sets of captions. Something different happens with BERTScore, its ability to take into account the context of the words makes it capable of identifying the grammatical correctness. 

We can see that the intersection of the \emph{Shuffle all} and the original captions is very small, also, the bell curves of the \emph{Shuffle $50\%$} and \emph{Shuffle $25\%$} are easy to distinguish. On the contrary, both the CLIPScore and CLIPScore$^{\text{ref}}$ are fooled by this exercise. The reason is that the CLIP embedding is taken from the output of the first token of the encoder, and even when it can \emph{perceive} the rest of the caption, the order of the words is not very important.

\subsection{Discussion on the metrics' findings}

The main goal of a scoring metric for IC, should be to clearly distinguish, between several models and which one of them is better than the other ones. In other words, from a some sets of different proposed natural language generated captions, which one of these sets is better than the other ones on an overall way, but also to be able to {\it order} how much better are these sets of proposed captions between each other.

According to our results, we found that $n$-gram based metrics are not always suitable to correctly assess this, on an artificially set of increasingly worse sets of captions. 
In Table \ref{tab:spearman}, we present a summary of the Spearman coefficient $(\rho_s)$ for all the analyzed metrics, using both methodologies.

For the {\it replacing} methodology, the CLIPScore$^{\text{ref}}$ achieved the best performance maintaining the known order, therefore it would be the best score under this single criterion. But we must remember that the caption should also be reported via Natural Language, hence, in a correctly grammatically written sentence. 

For that, we applied the {\it shuffling} criterion, to rank the metrics on how well they perform under a sentence that includes elements in the image, but is written in a way that is not understandable by a human. In this second methodology, BERTScore was the undisputed winner. BERTScore seems to be better suited to report the {\it natural language} property of the sentence, while CLIPScore is better suited to report how well the sentence is including elements present in the image.

With these findings, it is not absurd to think about getting rid entirely of the $n$-gram based metrics, since better metrics are already present that perform way better than the so-called {\it classical} ones. The embedding based metrics such as BERTScore and CLIPScore already achieved a performance that is not comparable anymore with BLEU, METEOR, SPICE, \emph{etc.} 

We understand that there was the need (or maybe still is) for a period of transition between the use of {\it classical} metrics since new models still needed to be compared to older ones, and the total adoption of better scores that do not only look for words (or tokens or $n$-grams) can not be immediately implemented. But we also think that we are already in a place of the SoTA where new ways of scoring the captions are needed that also take into account grammar and context. BERTScore and CLIPScore seem to be the first contenders in this new generation of metrics, and $n$-gram based metrics should leave a way for these new methods.

\begin{table}[!ht]
    \centering
    \begin{tabular}{lcc}
\hline
Metric  &  Replacing &   Shuffling  \\
 & $\rho_{s}$ & $\rho_{s}$ \\ \hline
BLEU-1 \cite{Papineni02bleu:a} & 0.640800 & 0.027080\\
BLEU-2 \cite{Papineni02bleu:a} & 0.555879 & 0.380119\\
BLEU-3 \cite{Papineni02bleu:a} & 0.330240 & 0.338320 \\
BLEU-4 \cite{Papineni02bleu:a} & 0.128960 & 0.134680\\
SacreBLEU \cite{post-2018-call} & 0.560640 & 0.474120 \\
METEOR \cite{Lavie2007}  & 0.575520 & 0.216000 \\
SPICE \cite{anderson2016spice} & 0.661920 & 0.230680\\
CIDEr \cite{vedantam2015cider} & 0.708320 & 0.123240\\
ROUGE-L \cite{lin-och-2004-orange}  & 0.570840 & 0.171400\\
BERTScore \cite{zhang2019bertscore} & 0.556920 & \textbf{0.709639}\\
CLIPScore \cite{DBLP:journals/corr/abs-2103-00020} & 0.774080 & 0.225760\\
CLIPScore$^{\text{ref}}$ \cite{DBLP:journals/corr/abs-2103-00020} & \textbf{0.805640} & 0.280159\\

\hline
\end{tabular}
    \caption{Summary of the Spearman correlation coefficients ($\rho_s$) for the metrics presented, using the {\it replacing} and the {\it shuffling} methodology. Bold correlations indicate the best results.}
    \label{tab:spearman}
\end{table}


\subsection{Real models}

We applied our methodology to captions generated by real models and compare their results. We selected three captioning methods, which are Meshed-Memory Transformer\footnote{\url{https://github.com/aimagelab/meshed-memory-transformer}} ($\mathcal{M}^2$-Transformer) \cite{cornia2020m2}, ConvCap\footnote{\url{https://github.com/aditya12agd5/convcap}} \cite{AnejaConvImgCap17} and the Google's implementation of Show, Attend \& Tell\footnote{ \url{https://www.tensorflow.org/tutorials/text/image_captioning}} (SA\&T) \cite{pmlr-v37-xuc15}, the latter trained for 30 epochs and a vocabulary of 10k tokens. 

As a first comparison, we applied the models to the validation set of MS-COCO 2017 and obtained their scores for all the metrics aforementioned. The results are presented in Table \ref{tab:models}.

\begin{table}[!ht]
    
    \footnotesize
    \centering
\resizebox{\textwidth}{!}{
    \begin{tabular}{lcccccc}
\hline
\bf Model  & \bf BLEU-4 & \bf SacreBLEU & \bf METEOR & \bf ROUGE-L & \bf CIDEr & \bf SPICE \\
\hline
SA\&T \cite{pmlr-v37-xuc15} 
& 0.0860 & 0.0781 & 0.1573 & 0.3403 & 0.3237 & 0.1029 \\
ConvCap \cite{AnejaConvImgCap17}
& 0.3658 & 0.3091 & 0.2790 & 0.5747 & 1.1727 & 0.2087 \\
$\mathcal{M}^2$-Transf \cite{cornia2020m2} 
& 0.3909 & 0.3399 & 0.2918 & 0.5863 & 1.3120 & 0.2259 \\

\hline
\\
\hline
\bf Model  & \bf BERT & \bf CLIP &  \bf CLIP$^\text{ref}$\\
\hline
SA\&T \cite{pmlr-v37-xuc15} 
& 0.9043 & 0.6497 & 0.6995\\
ConvCap \cite{AnejaConvImgCap17}
& 0.9403 & 0.7178 & 0.7865\\
$\mathcal{M}^2$-Transf \cite{cornia2020m2} 
& 0.9375 & 0.7344 & 0.7925\\

\hline

\end{tabular}
}
    \caption{Summary of the metrics evaluated for the validation set of MS-COCO 2017, for three well-known IC models.}
    \label{tab:models}
\end{table}

Here, we can see a big difference between SA\&T and the other two models' results in BLEU-4, SacreBLEU, METEOR, ROUGE-L, CIDEr, and SPICE but the scores are tighter in BERTScore and CLIPScore. The captions produced by $\mathcal{M}^2$-Transformer got higher results on all the metrics but in BERTScore. 

To find an explanation, we visually inspected its captions and found many of them look incomplete. It appears to be an effect of the Reinforcement Learning with the CIDEr metric as a reward, used during the $\mathcal{M}^2$-Transformer training. It is important to note that this model is ranked number one in CIDEr worldwide at the time of writing; it suggests an overfitting effect on the training.

The above is an example of how higher scores do not always mean better-generated captions. Since the beginning of IC, MS-COCO has been the most popular set for training and testing using the same metrics (BLEU, METEOR, ROUGE-L, and CIDEr), but very little has changed. We think changing and adapting the captioning metrics is essential to measure the method's results precisely.

In the case of the Reinforcement Learning step, we believe that if the community does not transition to better metrics, we will continue creating models that are specialists in fooling themselves, only getting the best result and highest metrics but not necessarily producing better quality captions. 


As a final experiment, we compare the three real models and the human-generated captions. For this, we used the metrics BERTScore, CLIPScore and CLIPScore$^{\text{ref}}$, since they got better results in Section \ref{sec:ranking}.
As we did in prior experiments, from the MS-COCO validation set, we took the first caption in the references as the Human set of captions, and the other four as the new set of references. For each real model, we formed a new set of captions with the union of the captions of that model and the set of Human captions, in that order. 

Next, we applied the metrics to this set and re-ordered the captions using the metric scores. Finally, we measure the order of the original caption by re-ordering of the metrics using the Spearman correlation. We hypothesized that the human-generated captions were better than the captions generated by the real models, and we could measure it with the Spearman correlation.

\begin{figure}
  \centering
 
  \subfloat[BERTScore $(\rho_s= 0.5132)$]{\includegraphics[width=0.5\textwidth]{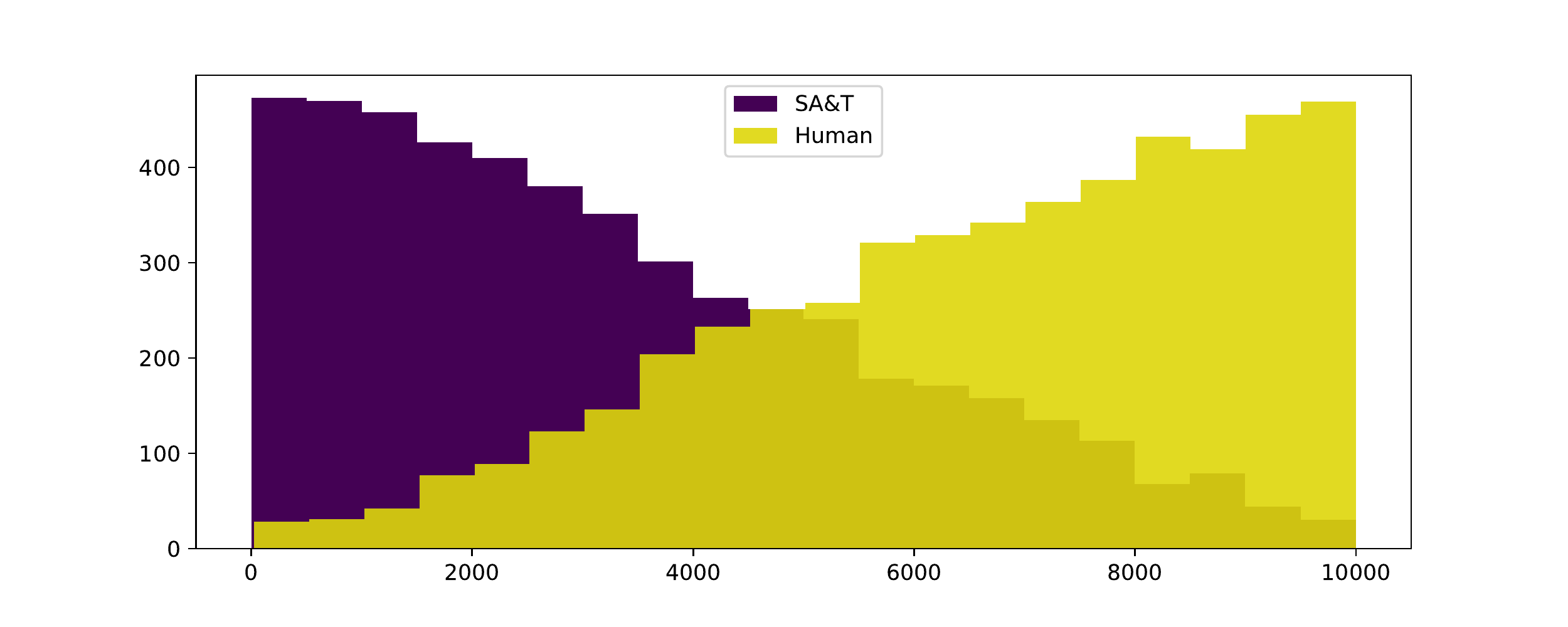}}\hfill
  \subfloat[CLIPScore $(\rho_s= 0.3952)$]{\includegraphics[width=0.5\textwidth]{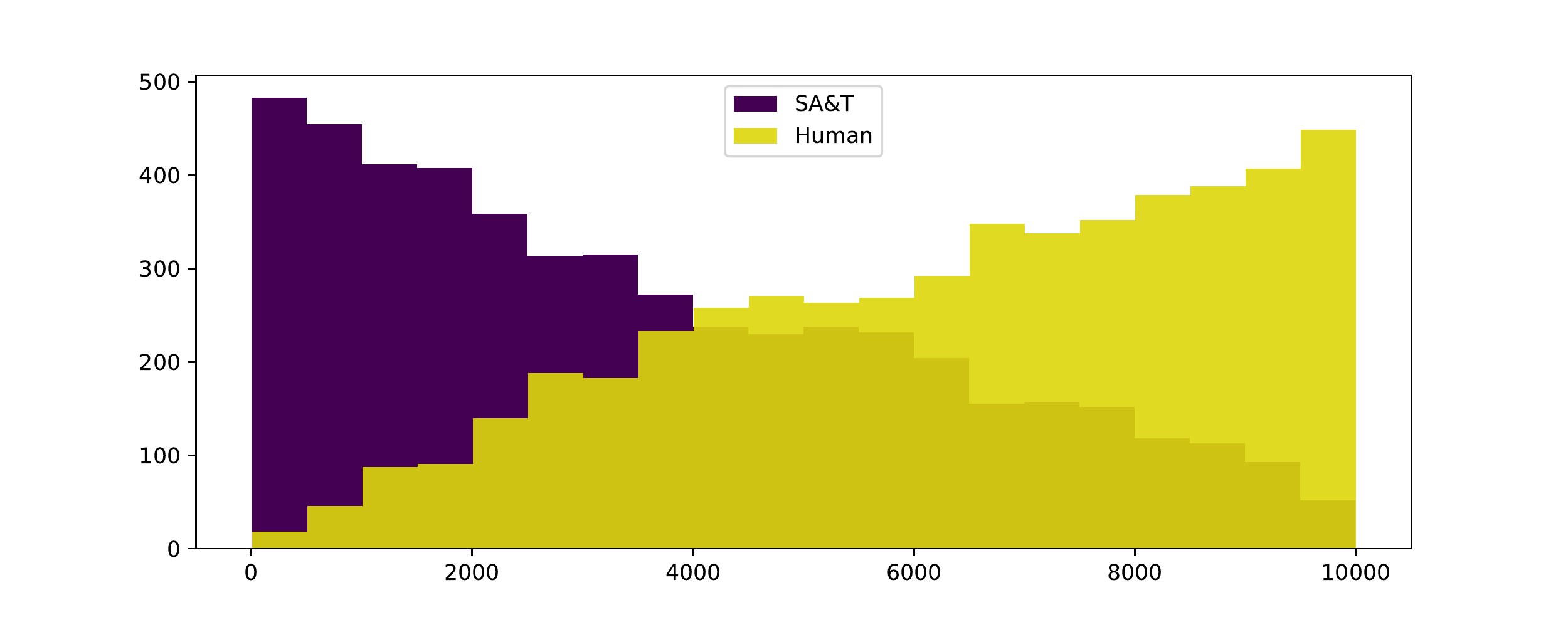}}
  \subfloat[CLIPScore$^{\text{ref}}$ $(\rho_s=
  0.4816)$]{\includegraphics[width=0.5\textwidth]{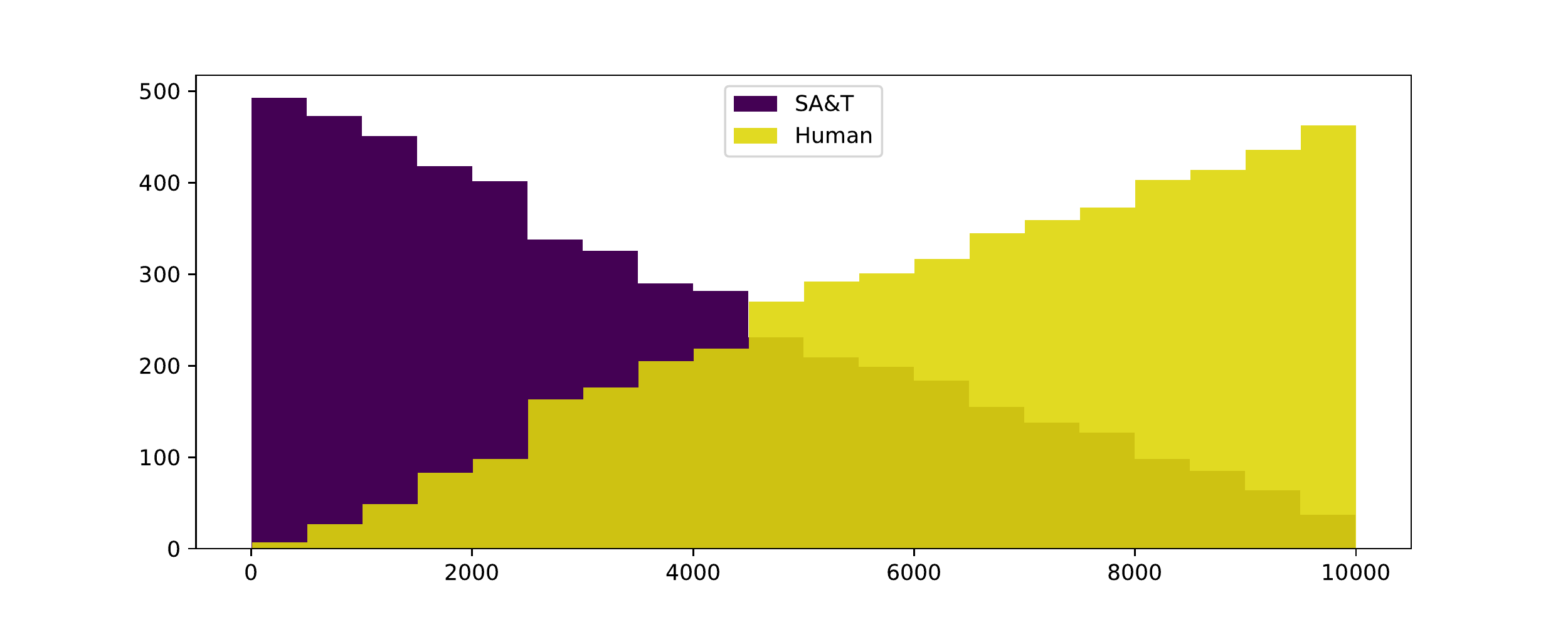}}
  \caption{Show Attend and Tell comparison with humans.\label{fig:saat-vs-humans}}
\end{figure}

In Figure~\ref{fig:saat-vs-humans} we show the histograms of the re-ordering of the SA\&T model. Here,  we can see that the three metrics do a good job telling apart the captions from SA\&T and from the humans. So, we can say that the captions of this model are not close in quality to those of humans. 

\begin{figure}
  \centering
 
  \subfloat[BERTScore $(\rho_s= -0.0376)$]{\includegraphics[width=0.5\textwidth]{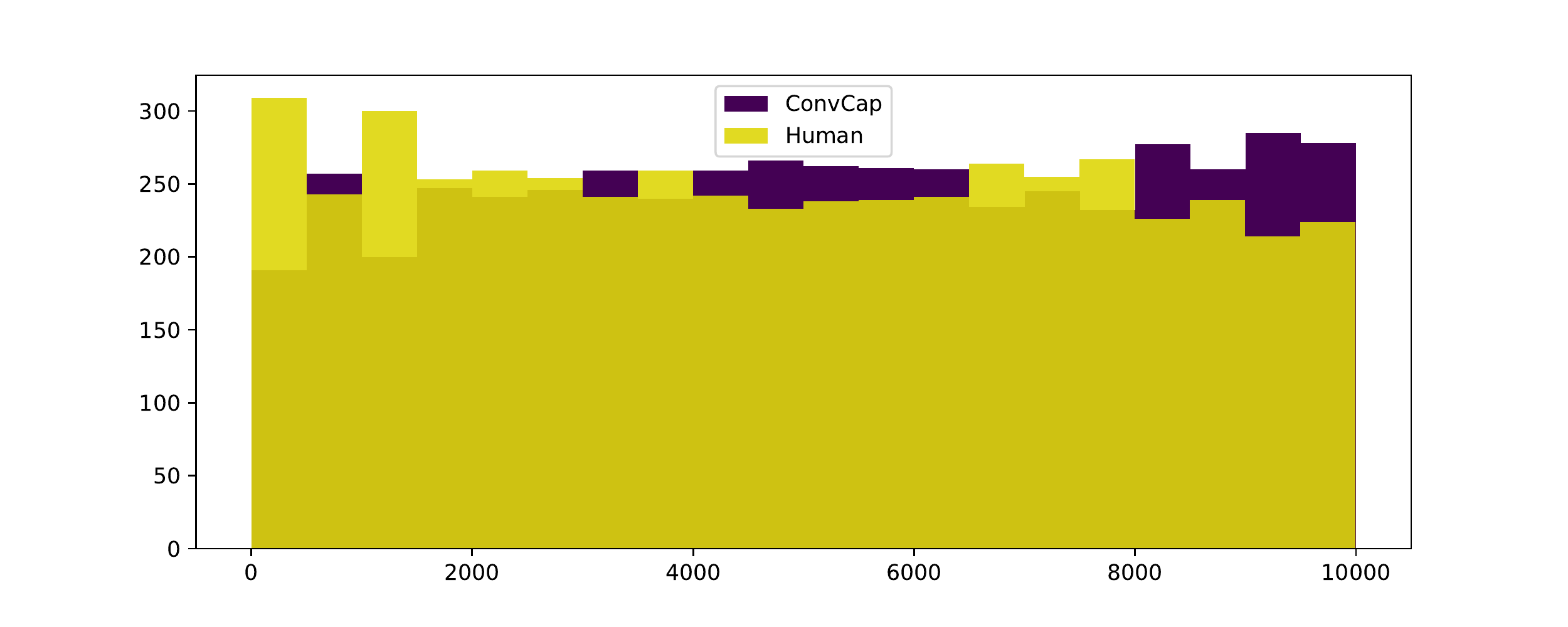}}\hfill
  \subfloat[CLIPScore $(\rho_s= 0.1708)$]{\includegraphics[width=0.5\textwidth]{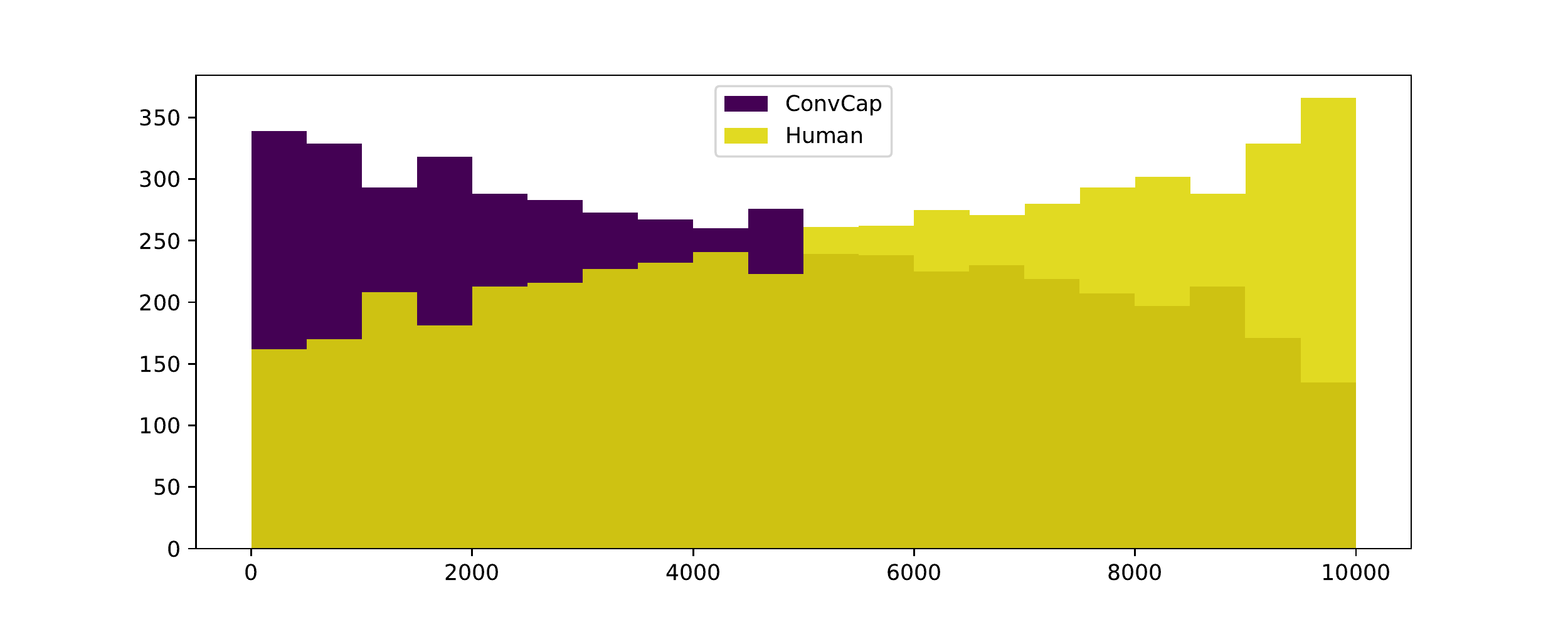}}
  \subfloat[CLIPScore$^{\text{ref}}$ $(\rho_s=
  0.0824)$]{\includegraphics[width=0.5\textwidth]{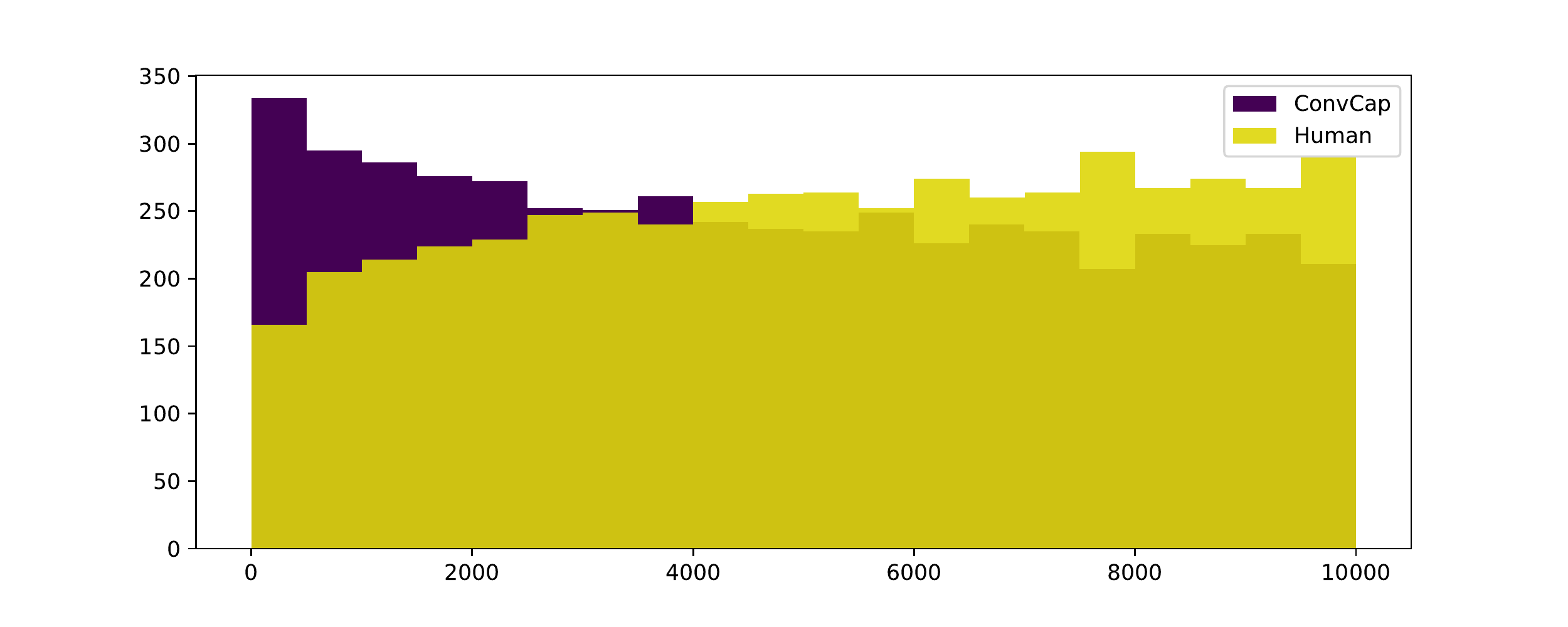}}
  \caption{ConvCap comparison with humans.\label{fig:convcap-vs-humans}}
\end{figure}

Let us focus now on the ConvCap model. In Figure~\ref{fig:convcap-vs-humans} we show its histograms. Here, something unexpected happened; the captions of ConvCap got better BERTScore results than humans; this is reflected because of the negative Spearman correlation. The histograms of CLIPScore and CLIPScore$^{\text{ref}}$ show that the human captions have a slight advantage. We can say that the ConvCap captions are close to humans in quality.

\begin{figure}
  \centering
 
  \subfloat[BERTScore $(\rho_s= 0.0016)$]{\includegraphics[width=0.5\textwidth]{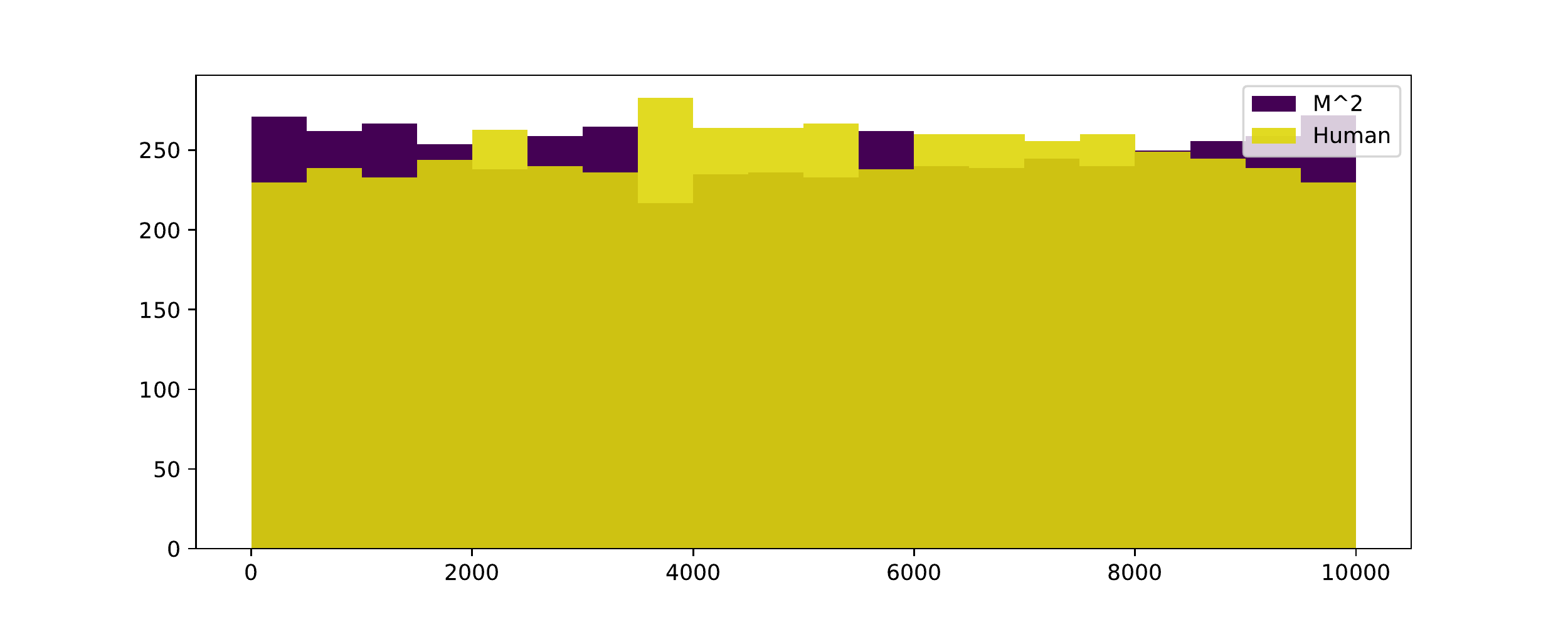}}\hfill
  \subfloat[CLIPScore $(\rho_s= 0.0860)$]{\includegraphics[width=0.5\textwidth]{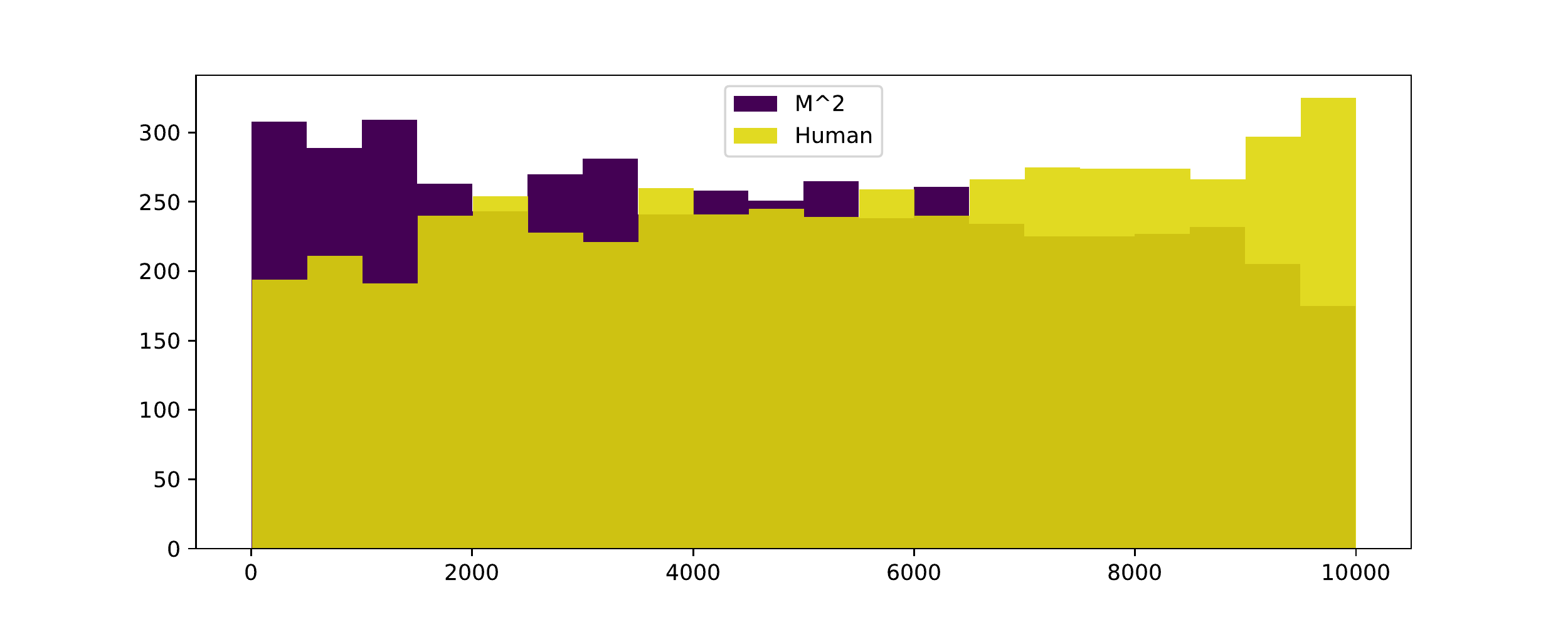}}
  \subfloat[CLIPScore$^{\text{ref}}$ $(\rho_s=
  0.0452)$]{\includegraphics[width=0.5\textwidth]{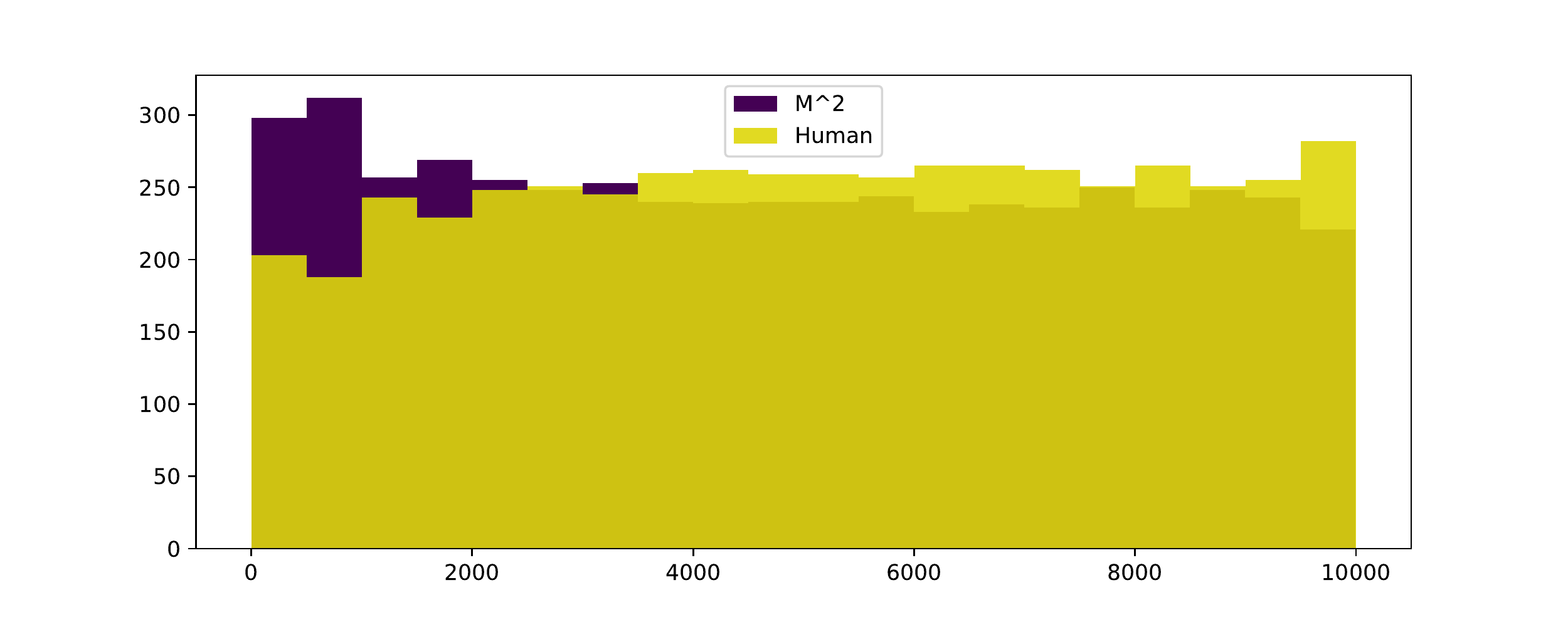}}
  \caption{$\mathcal{M}^2$-Transformer comparison with humans.\label{fig:M2-vs-humans}}
\end{figure}

Finally, the $\mathcal{M}^2$-Transformer histograms are shown in Figure~\ref{fig:M2-vs-humans}. The Spearman correlations are very low in all the metrics, which implies that the captions from the model are indistinguishable from those produced by humans. So we then have two possibilities; the IC models are now at the same level as humans or we need more sophisticated metrics that could tell us what is the \emph{real} difference between the two sets of captions.

\section{Conclusions}\label{sec:conclusions}
In this work, we present an evaluation and comparison of a set of well-known evaluation metrics for the Image Captioning task. We designed two kinds of experiments, generating captions with different qualities and properties. We then tested the metrics to measure how they can identify and separate the captions. Also, we used some models from state-of-the-art to apply our methodology with real captions. 

From this evaluation and comparison, some findings were found. In the case of $n$-gram based metrics, we observed that they are not adequate to assess these sets of worse and better captions. BERTScore prefers grammar correctness over caption accuracy when it is presented as a candidate caption, even though the candidate sentence does not correspond to the references sentences. On the other hand, CLIPScore excels at identifying the words that describe elements present in the image but pays little attention to the grammar. Our experiments with captions produced by real models suggest that it is dangerous to continue the evaluation using metrics like BLEU and CIDEr and we might be overlooking problems like overfitting.

On the other hand, we are getting to the point where models are getting comparable to human capabilities, most of the metrics presented in this paper use the human-generated references as the \emph{best} possible captions. We must start thinking of new ways of comparing which one is the \emph{best} possible caption for a certain image. CLIPScore is a clear candidate to be the prototype of a new generation of metrics that do not take a human annotation as the indisputable best option description of an image. Otherwise, we foresee that models will begin to resemble an \emph{asymptotic} behavior in which they will try to behave closer to humans, not necessarily meaning that they get better in the Image Captioning task.

In conclusion, we think it is necessary to pay less attention to the $n$-gram based metrics since more complete alternatives are already being used and perform way better. The embedding-based metrics such as BERTScore and CLIPScore achieved better performance, and they are not comparable anymore with BLEU, METEOR, SPICE, {\it etc}. The research community for IC tasks should transition to novel ways of evaluating the captions generated by their models.

\section*{Acknowledgments}
This work has been done through CONACYT (National Council of Science and Technology from Mexico) support with Ciencia Básica grant with project ID A1-S-34811.

\bibliographystyle{elsarticle-num-names}
\bibliography{sn-bibliography}

\end{document}